\documentclass[letterpaper]{article} 
\usepackage[preprint]{aaai2027}  
\usepackage[hyphens]{url}  
\usepackage{graphicx} 
\usepackage{natbib}  
\usepackage{caption} 
\usepackage{algorithm}
\usepackage{algorithmic}

\usepackage{newfloat}
\usepackage{listings}
\DeclareCaptionStyle{ruled}{labelfont=normalfont,labelsep=colon,strut=off} 
\floatstyle{ruled}
\newfloat{listing}{tb}{lst}{}
\floatname{listing}{Listing}

\usepackage{booktabs}

\usepackage{tabularx}
\usepackage{subcaption}
\usepackage{colortbl}
\usepackage{enumitem}
\usepackage{xcolor}
\usepackage{amsmath}
\usepackage{amssymb}
\usepackage{amsfonts}

\newcommand{\corresponding}{\textsuperscript{*}}
\title{Native Multilingual Chain-of-Thought Reasoning in Low-Resource \\Southeast Asian Languages}  
\author{
    Sean Gip Lim\textsuperscript{\rm 1,2,3}\corresponding,
    William Chandra Tjhi\textsuperscript{\rm 2},
    Hai Leong Chieu\textsuperscript{\rm 3}
}
\affiliations{
    \textsuperscript{\rm 1}Nanyang Technological University\\
    \textsuperscript{\rm 2}AI Singapore\\
    \textsuperscript{\rm 3}DSO National Laboratories\\
    
    li0002ip@e.ntu.edu.sg, wtjhi@aisingapore.org, chaileon@dso.org.sg
}

\begin{document}

\maketitle

\begin{abstract}
Large Language Models have achieved substantial progress in reasoning capabilities.
Yet in low-resource native settings, many suffer from cross-lingual collapse, 
reverting to English during intermediate steps that require complex logical reasoning.
This presents a cold-start bottleneck for policy optimization,
whereas standard fine-tuning risks catastrophic forgetting due to cross-lingual representation drift.
To address these challenges, we introduce the Onramp-Sequence Cross-Distillation (OSCD),
a post-training algorithm that projects high-resource reasoning trajectories into low-resource vocabulary subspaces during generative training rollouts via an integrated translator agentic loop, 
ensuring the stable and efficient translation of dynamically generated reference samples for fine-tuning.
This is coupled with joint-embedding semantic alignment of both reference and target-language reasoning traces,
thereby bridging the pairwise cross-lingual representational gaps.
Comprehensive evaluations using the \texttt{AIME25} and \texttt{HMMT25} benchmarks demonstrate that OSCD yields up to 3.2 times overall improvements in native Southeast Asian languages for mathematical reasoning, 
of which the joint-embedding semantic alignment component 
contributes up to 6.4\% improvements in linguistic debiasing over translation-only baselines.
\end{abstract}

\begin{links}
    \link{Code, Dataset, Model}{https://github.com/SG-Lim/OSCD}
\end{links}

\section{Introduction}
Large Language Models (LLMs) have achieved remarkable reasoning capabilities, 
largely driven by advancements in Chain-of-Thought (CoT) prompting and Reinforcement Learning (RL) (\citealp{3600270.3602070}; \citealp{shao2024deepseekmathpushinglimitsmathematical}). 
Nevertheless, these advancements remain overwhelmingly conditioned on high-resource typographic regimes, predominantly English 
(\citealp{tran2025reasoningtransferextremelylowresource}; \citealp{schut2025multilingualllmsthinkenglish}; \citealp{barua2026longchainofthoughtreasoninglanguages}). 
Consequently, linguistic regions characterized by severe data deficits are left behind.
Southeast Asia (SEA), for instance, 
represents a population of 671 million but suffers from a lack of training data for regional development of native frontier capabilities \citep{lovenia-etal-2024-seacrowd}.
To compensate for this resource asymmetry, multilingual models inherently exhibit an English-centric bias, 
leveraging high-resource languages as a structural anchor for low-resource problem-solving \citep{schut2025multilingualllmsthinkenglish}.
This creates a fundamental barrier to user accessibility, 
wherein the ability to interpret the step-by-step reasoning trace remains paramount to critical domains such as education and research.
Because the intermediate process of CoT generation reverts systemically to English following complex logical transitions (\citealp{park2026crosslingualcollapselanguagecentricfoundation}; \citealp{zhao2025comprehensiveevaluationmultilingualchainofthought}; \citealp{kang2026multilingualreasoninggapsemerge}),
non-English native speakers are placed at a severe disadvantage, 
undermining the true value and equity of artificial intelligence.

\begin{figure}[t]
\centering
\includegraphics[width=0.95\columnwidth]{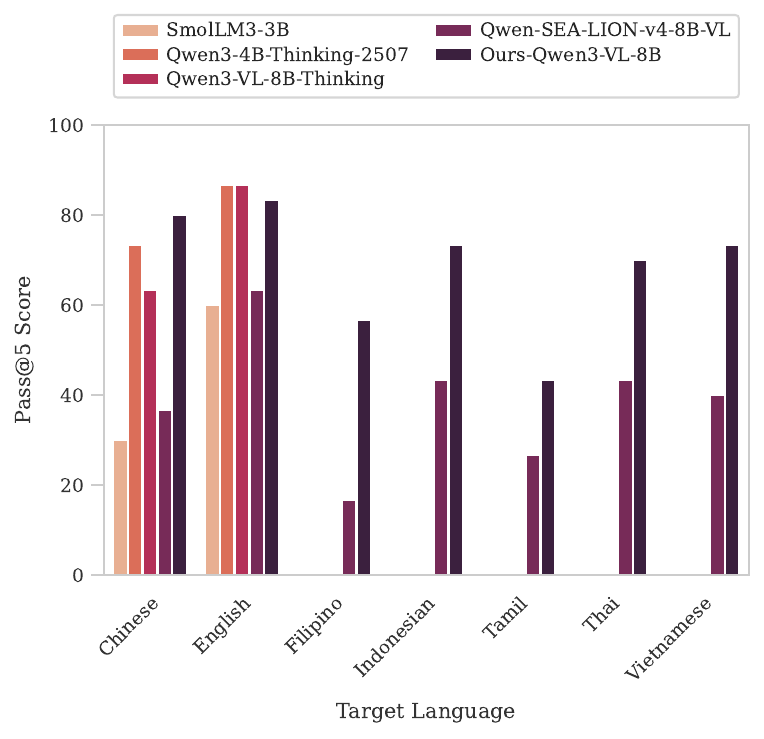} 
\caption{\texttt{Pass@5} evaluation of large language models on the \texttt{AIME25} benchmark, with linguistic alignment verifications across 7 target languages (\texttt{ZH,EN,Fi,IN,TA,TH,VI}).}
\label{bar_plots}
\end{figure}

\begin{figure*}[t]
\centering
\includegraphics[width=0.85\textwidth]{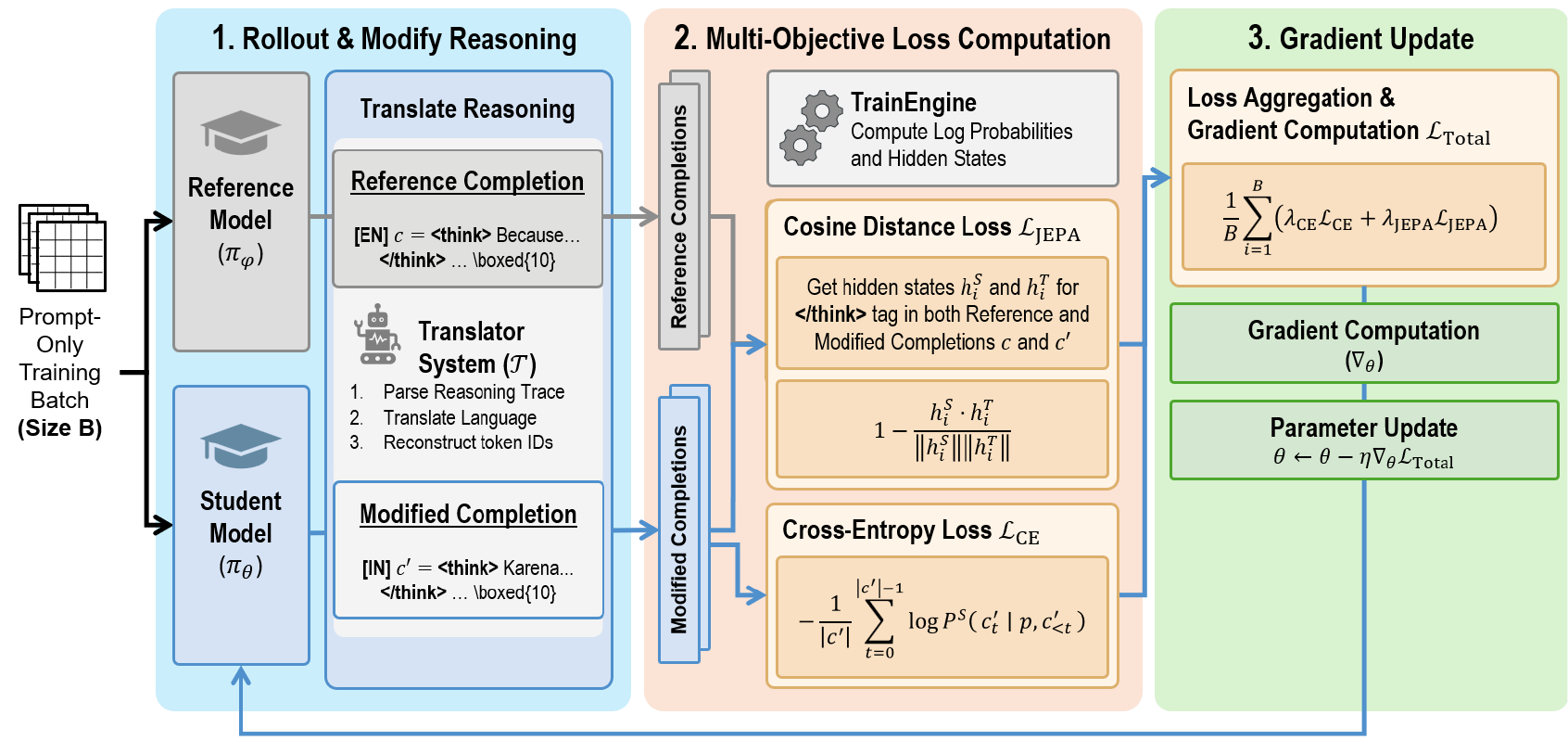} 
\caption{Graphical representation of the Onramp Sequence Cross-Distillation (OSCD) algorithm.}
\label{algorithm_OSCD}
\end{figure*}

\begin{figure}[t]
\centering
\includegraphics[width=0.85\columnwidth]{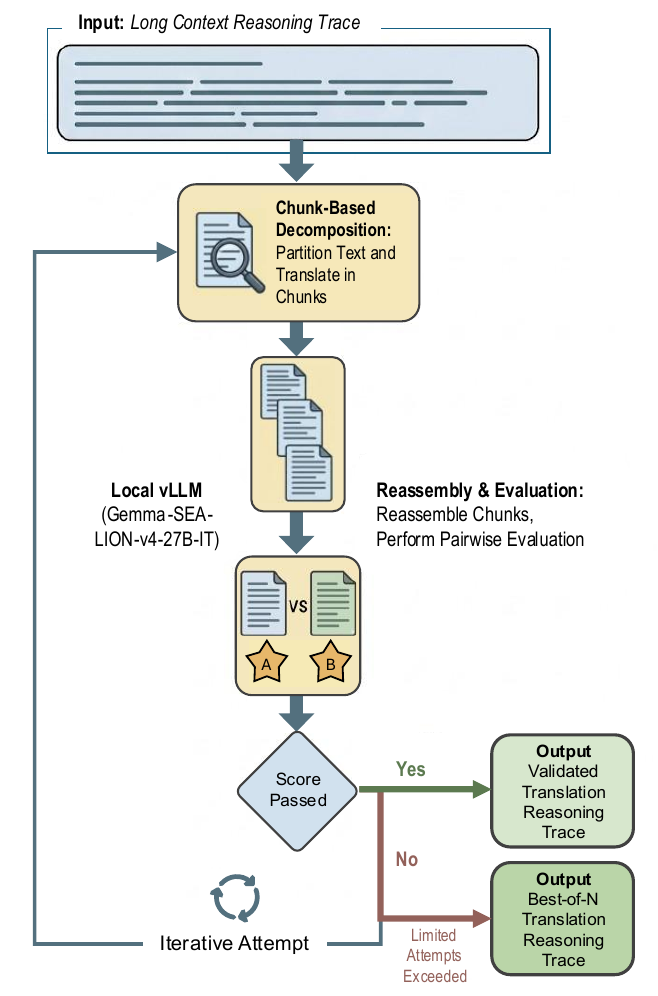} 
\caption{Graphical representation of the agentic loop system for translation of long-context reasoning traces.}
\label{agentic_translator}
\end{figure}

Because many thinking-enabled models inherently struggle to generate low-resource native trajectories on complex, multi-step reasoning tasks (Figure~\ref{bar_plots}),
we identify a critical bottleneck when applying modern RL approaches, 
particularly policy optimization algorithms relying on pre-existing knowledge space and reasoning trajectories to optimize sampling efficiency (\citealp{chen2026does}; \citealp{kim2025rlvr}). 
This presents a cold-start training problem,
wherein absent of successful native samples to trigger sufficient rewards, the model cannot optimize its policy away from high-resource English pathways.
On the other hand, standard Supervised Fine-Tuning (SFT) or Continued Pre-Training (CPT) on curated multilingual datasets introduces separate challenges,
most notably the risk of catastrophic forgetting (\citealp{alexandrov-etal-2024-mitigating}; \citealp{aggarwal-etal-2024-maple}; \citealp{liu-niehues-2025-conditions}).
This is largely attributed to the per-language misalignment of model hidden state representations,
causing the mapping of parallel texts with linguistic differences to disjoint embedding subspaces despite semantic similarities (\citealp{aggarwal2024exploring}; \citealp{lim2025languagespecificlatentprocesshinders}; \citealp{10.1609/aaai.v39i27.35038}). 
As a result, naive training without bridging this gap exacerbates representation drift (\citealp{gurgurov2026sparsesubnetworkenhancementunderrepresented}), 
risking instead the established geometry of high-resource embeddings.

To address both the representational bottleneck and cold-start RL dilemma, 
we propose a training framework that optimizes for semantic equivalence of generative rollouts across different languages, 
while expanding the knowledge search space for low-resource multilingual reasoning (Figure~\ref{algorithm_OSCD}). 
Our summary of key contributions are as follows:
\begin{itemize}
    \item We introduce OSCD, a novel training framework designed to enable native multilingual CoT reasoning in low-resource regional languages, specifically SEA languages. This involves the dynamic, cross-policy localization of generative rollouts for fine-tuning, coupled with joint-embedding semantic alignment bridging the representational differences of parallel texts across different languages.
    \item Extensive experiments on open-ended, mathematical reasoning tasks demonstrate the efficiency and robustness of our training approach. We prove that the reasoning capabilities of English-dominant models can be effectively transferred to low-resource settings using synthetically localized data, therefore expanding the multilingual knowledge search space required for subsequent RL.
    \item Comprehensive evaluations on the \texttt{AIME25} and \texttt{HMMT25} benchmarks across language variants show that models post-trained with OSCD outperform existing multilingual SEA model of a comparable scale. Our training framework preserves model intelligence in high-resource settings, while successfully addressing systemic linguistic biases that induce unwanted language fallbacks. 
\end{itemize}

\section{Related Work}
\subsubsection{Multilingual Reasoning}
Multilingual reasoning is a major NLP challenge, particularly for low-resource languages \citep{tran2025reasoningtransferextremelylowresource}.
Consequently, models often map decisions into an English-adjacent latent space before generating target-language outputs \citep{schut2025multilingualllmsthinkenglish}. 
This English-centric bias causes cross-lingual understanding failures, uneven reasoning quality, and 'cross-lingual collapse' where intermediate CoT reverts to English under increased difficulty (\citealp{hwang2025learngloballyspeaklocally}; \citealp{park2026crosslingualcollapselanguagecentricfoundation}; \citealp{zhao2025comprehensiveevaluationmultilingualchainofthought}; \citealp{kang2026multilingualreasoninggapsemerge}).

To bridge this gap, several methods leverage English as an intermediary via translation, code-switching, cross-lingual distillation, or parallel fine-tuning (\citealp{chen2024breakinglanguagebarriersmultilingual2}; \citealp{kang2026multilingualreasoninggapsemerge}; \citealp{chai2024xcotcrosslingualinstructiontuning2}; \citealp{barua2026longchainofthoughtreasoninglanguages}; \citealp{zheng2026adamcotrethinkingcrosslingualfactual}). 
Native reasoning approaches include mapping problems into language-agnostic symbolic spaces (\citealp{ranaldi-pucci-2025-multilingual}),
causal interventions to subtract language-specific hidden states (\citealp{zhao2025languagemorelanguagereasoningdisentanglement}), 
and using reinforcement learning for language-consistency rewards (\citealp{hwang2025learngloballyspeaklocally}). 
Nevertheless, native target-language reasoning systematically lags English-pivoted approaches, suffering from language-specific generation errors and conceptual misunderstandings (\citealp{barua2026longchainofthoughtreasoninglanguages}).

\subsubsection{Joint-Embedding Predictive Architecture}
The mapping of varying semantic contexts into a shared latent space aligns with the foundational principles of Joint-Embedding Predictive Architecture (JEPA) (\citealp{garrido2024learningleveragingworldmodels}; \citealp{maes2026leworldmodelstableendtoendjointembedding}).
\citet{huang2025llmjepalargelanguagemodels} introduced LLM-JEPA, combining standard autoregression with an embedding-space prediction loss to align different semantic views of the same knowledge.
Subsequent frameworks leverage this objective to bound hidden-state trajectories via geometric regularizers (\citealp{huang2026semantictubepredictionbeating}; \citealp{yuan2026semanticsteppredictionmultistep}), 
decouple latent reasoning from token generation (\citealp{liu2026jepareasonerdecouplinglatentreasoning}), 
and project query and document embeddings into shared spaces (\citealp{Chen_2026}).

\citet{lim2025languagespecificlatentprocesshinders} demonstrated the importance of aligning semantic properties in low-resource multilingual contexts, without which models typically default to disjoint, less accurate representations.
Along the topic, contrastive learning applied to monolingual English data have been shown to project cross-lingual representations into a shared, language-invariant space (\citealp{wang-etal-2022-english}),
whereas minimizing cross-lingual divergence between probabilistic latent variables guides models to capture structured, language-agnostic semantic representations (\citealp{sherborne-etal-2023-optimal}).

\section{Methodology}
The OSCD framework comprises two components: 
(1) Rollout with Sequence Mutation, of which dynamically synthesizes native-language reasoning trajectories to populate low-resource vocabulary subspaces,
and (2) Multi-Objective Cross-Distillation, which enforces sequence-level semantic alignment while fine-tuning is carried out on completions with localized reasoning traces.

\subsection{Rollout with Sequence Mutation}
Given a prompt $p$ targeting a low-resource language $l \in \mathcal{L}$, 
the pipeline initializes by sampling a completion sequence $c \sim \pi_\phi(\cdot \mid p)$ from a high-resource reference model $\pi_\phi$. 
To isolate its intermediate CoT, we decode the token sequence into a text string $s = \mathcal{D}(c)$, where $\mathcal{D}(\cdot)$ is the decoding function. 
Using a predefined set of structural delimiters $\Delta = \{\delta_{\text{open}}, \delta_{\text{close}}\}$ (e.g., opening and closing think tags), 
we partition $s$ into a reasoning string $s_{r}$ and an answer string $s_{a}$:
\begin{equation}
    (s_{r}, s_{a}) = \text{Split}\big(\mathcal{D}(c), \Delta\big)
\end{equation}

To localize dynamically the reasoning process into the low-resource vocabulary subspace $\mathcal{V}_{L}$, 
an external translation function $\mathcal{T}$ is then applied to the generated reasoning string,
yielding a low-resource native trace conditioned on $l$:
\begin{equation}
    s'_{r} = \mathcal{T}(s_{r}, l)
\end{equation}

Thereafter,
the text segments are projected back into the discrete token space while preserving topological boundaries of the original completion.
Let $\mathcal{E}^*(\cdot)$ denote the controlled encoding function. 
The mutated completion sequence $c'$ is reconstructed via an ordered concatenation ($\oplus$) of the re-encoded segments and their structural delimiters:
\begin{equation}
    c' = \delta_{\text{open}} \oplus \mathcal{E}^*(s'_r) \oplus \delta_{\text{close}} \oplus \mathcal{E}^*(s_a)
\end{equation}

We constrain the translation function $\mathcal{T}$ entirely to the reasoning trace, 
preserving the teacher's original answer string $s_a$ verbatim. 
This aims to establish a stable anchor, 
thus guaranteeing that the underlying inferential trajectory remains bound to a known, valid outcome.

\subsection{Multi-Objective Cross-Distillation}
\subsubsection{Full-Sequence Cross-Entropy}
To drive native multilingual acquisition, 
we apply full-sequence log-likelihood ($\mathcal{L}_{\text{CE}}$) across the entirety of the student completion,
initializing the student model $\pi_\theta$ as its reference teacher $\pi_\phi$.
This ensures $\pi_\theta$ constructs its intermediate reasoning steps over the target medium $\mathcal{V}_{L}$,
while remaining anchored to a stable answer in close proximity to its original distribution:
\begin{equation}
    \mathcal{L}_{\text{CE}} = - \frac{1}{|c'|} \sum_{t=0}^{|c'|-1} \log \pi_\theta(c'_{t} \mid p, c'_{<t})
\end{equation}

\subsubsection{Joint-Embedding Semantic Alignment}
While $\mathcal{L}_{\text{CE}}$ induces a language shift, naively fine-tuning on translated text risks further isolation and drifting of model representations into language-specific subspaces \citep{lim2025languagespecificlatentprocesshinders}.
To avert this concern, we introduce a secondary objective $\mathcal{L}_{\text{JEPA}}$, 
referencing and adapting from \citet{huang2025llmjepalargelanguagemodels} to fulfil our goal of 
bridging representational differences through latent-space alignment of specific pairwise tokens.

Let $h^T_k$ and $h^S_{k'}$ denote the last-layer hidden states corresponding to the closing think tag ($\delta_{\text{close}}$) at token indices $k$ and $k'$, respectively,
conditioned on native reasoning trace $r'$ for the student model $\pi_\theta$ and $r$ for teacher model $\pi_\phi$.
We minimize the cosine distance between these corresponding hidden states as follows:
\begin{equation}
    \mathcal{L}_{\text{JEPA}} = 1 - \frac{h^S_{k'} \cdot h^T_k}{\|h^S_{k'}\| \|h^T_k\|}
\end{equation}

We specifically target $\delta_{\text{close}}$, because it serves as an informational bottleneck encapsulating the aggregated semantics of reasoning process prior to answer generation,
while preserving token-level flexibility across different languages so the model is not overconstrained when generatively navigating its internal representations.

Consequently, the total loss is formalized as a multi-objective function:
\begin{equation}
    \mathcal{L}_{\text{total}} = \lambda_{\text{CE}} \mathcal{L}_{\text{CE}} + \lambda_{\text{JEPA}} \mathcal{L}_{\text{JEPA}}
\end{equation}

where $\lambda_{\text{CE}}$ and $\lambda_{\text{JEPA}}$ are hyperparameter weights corresponding to vocabulary acquisition and cross-lingual semantic alignment, respectively.

For post-training optimization, we train models for $1$ epoch in BF16 precision using the AdamW optimizer with a cosine scheduled learning rate of $2 \times 10^{-5}$, 
a $0.1$ warmup fraction, and a $1.0$ gradient clip norm. 
Training sequences are capped at $4096$ prompt tokens and $8192$ completion tokens. 
We also keep inference parameters for generative rollouts and reasoning benchmarks consistent with the default recommendations of the respective model developers, 
using a maximum completion length of $81920$ tokens for evaluations.
Both $\lambda_{\text{CE}}$ and $\lambda_{\text{JEPA}}$ are set to $1.0$ for simplicity of this study.

\begin{table*}[t]
\centering
\small
\setlength{\tabcolsep}{3pt}
\begin{tabular*}{\textwidth}{@{\extracolsep{\fill}}l c c c c c c c c c c}
\hline
 & \multicolumn{5}{c}{\textbf{AIME25}} & \multicolumn{5}{c}{\textbf{HMMT25}} \\
\cline{2-6} \cline{7-11}
 & \multicolumn{2}{c}{\textbf{Any-CoT}} & \multicolumn{3}{c}{\textbf{Target-CoT}} & \multicolumn{2}{c}{\textbf{Any-CoT}} & \multicolumn{3}{c}{\textbf{Target-CoT}} \\
\cline{2-3} \cline{4-6} \cline{7-8} \cline{9-11}
\textbf{Model} & \textbf{Pass@5} & \textbf{Mean@5} & \textbf{Pass@5} & \textbf{Mean@5} & \textbf{LRI} & \textbf{Pass@5} & \textbf{Mean@5} & \textbf{Pass@5} & \textbf{Mean@5} & \textbf{LRI} \\
\hline
Qwen-SEA-LION-v4-8B-VL & 55.6 {\scalebox{0.8}{$\pm$6.9}} & 37.3 {\scalebox{0.8}{$\pm$3.1}} & 47.8 {\scalebox{0.8}{$\pm$13.9}} & 28.4 {\scalebox{0.8}{$\pm$9.7}} & 75.2 & 37.8 {\scalebox{0.8}{$\pm$6.9}} & 24.2 {\scalebox{0.8}{$\pm$3.7}} & 32.2 {\scalebox{0.8}{$\pm$8.4}} & 17.8 {\scalebox{0.8}{$\pm$4.8}} & 74.2 \\
Qwen3-VL-8B-Thinking & \textbf{\color{red}82.2 {\scalebox{0.8}{$\pm$7.7}}} & \textbf{\color{red}69.6 {\scalebox{0.8}{$\pm$8.7}}} & 50.0 {\scalebox{0.8}{$\pm$44.8}} & 41.6 {\scalebox{0.8}{$\pm$38.0}} & 56.1 & \textbf{\color{red}52.2 {\scalebox{0.8}{$\pm$7.7}}} & \textbf{\color{red}44.4 {\scalebox{0.8}{$\pm$6.0}}} & 32.2 {\scalebox{0.8}{$\pm$29.1}} & 26.0 {\scalebox{0.8}{$\pm$25.1}} & 55.6 \\
\cellcolor{white!15}\textbf{Ours-Qwen3-8B-$\mathcal{L}_{\text{CE}}$} & 56.7 {\scalebox{0.8}{$\pm$14.5}} & 36.9 {\scalebox{0.8}{$\pm$16.0}} & 37.8 {\scalebox{0.8}{$\pm$31.0}} & 23.3 {\scalebox{0.8}{$\pm$24.9}} & 57.6 & 42.2 {\scalebox{0.8}{$\pm$15.4}} & 27.3 {\scalebox{0.8}{$\pm$10.5}} & 31.1 {\scalebox{0.8}{$\pm$22.7}} & 19.2 {\scalebox{0.8}{$\pm$16.7}} & 56.4 \\
\cellcolor{white!15}\textbf{Ours-Qwen3-8B-$\mathcal{L}_{\text{CE+JEPA}}$} & 58.9 {\scalebox{0.8}{$\pm$13.9}} & 36.0 {\scalebox{0.8}{$\pm$16.2}} & 40.0 {\scalebox{0.8}{$\pm$26.0}} & 24.2 {\scalebox{0.8}{$\pm$26.7}} & 59.4 & 35.6 {\scalebox{0.8}{$\pm$10.2}} & 21.1 {\scalebox{0.8}{$\pm$9.4}} & 27.8 {\scalebox{0.8}{$\pm$17.1}} & 15.0 {\scalebox{0.8}{$\pm$15.4}} & 57.0 \\
\cellcolor{white!15}\textbf{Ours-Qwen3-8B-$\mathcal{L}_{\text{CE (Agentic)}}$} & 80.0 {\scalebox{0.8}{$\pm$3.3}} & 62.7 {\scalebox{0.8}{$\pm$8.7}} & 74.4 {\scalebox{0.8}{$\pm$6.9}} & 52.8 {\scalebox{0.8}{$\pm$13.7}} & 85.4 & 50.0 {\scalebox{0.8}{$\pm$6.7}} & 37.6 {\scalebox{0.8}{$\pm$5.4}} & 47.8 {\scalebox{0.8}{$\pm$8.4}} & 28.3 {\scalebox{0.8}{$\pm$10.9}} & 81.1 \\
\cellcolor{white!15}\textbf{Ours-Qwen3-8B-$\mathcal{L}_{\text{CE+JEPA (Agentic)}}$} & 80.0 {\scalebox{0.8}{$\pm$3.3}} & 64.4 {\scalebox{0.8}{$\pm$5.7}} & \textbf{\color{red}78.9 {\scalebox{0.8}{$\pm$3.8}}} & \textbf{\color{red}54.2 {\scalebox{0.8}{$\pm$11.6}}} & \textbf{\color{red}87.9} & 51.1 {\scalebox{0.8}{$\pm$5.1}} & 39.3 {\scalebox{0.8}{$\pm$5.3}} & \textbf{\color{red}48.9 {\scalebox{0.8}{$\pm$5.1}}} & \textbf{\color{red}31.9 {\scalebox{0.8}{$\pm$9.1}}} & \textbf{\color{red}86.3} \\
\hline
\end{tabular*}
\caption{\label{overall-ablation-results}
Overall ablation performance comparison across \texttt{AIME25}, \texttt{HMMT25} benchmarks evaluating Any-CoT vs Target-CoT setups,  
averaged across 3 target languages (\texttt{ZH,EN,IN}) comparing different training loss configurations. 
}
\end{table*}

\begin{figure*}[t]  
    \centering
    \includegraphics[width=0.95\textwidth]{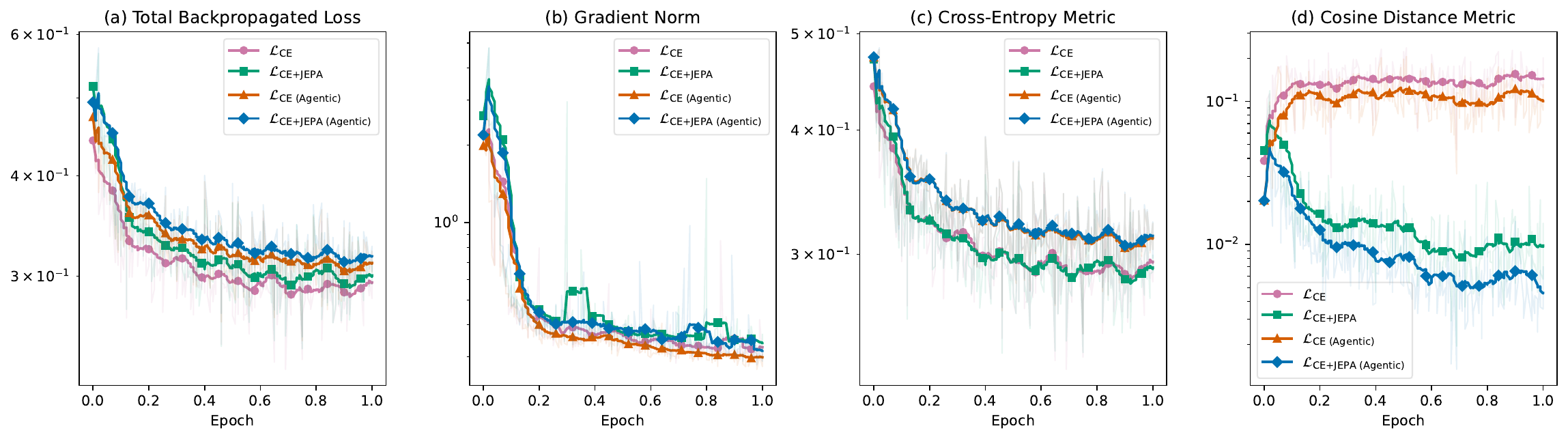} 
    \captionsetup{justification=centering}
    \caption{Training dynamics comparing different training loss configurations using \texttt{Qwen3-VL-8B-Thinking}. 
    }
    \label{ablation_curves}
\end{figure*}

\section{Experimentation}
The primary objective of this study is to empirically validate our proposed OSCD training framework.
To this end, we design our experiments to address the following core research questions:
\begin{enumerate}
    \item \textbf{Native Reasoning Accuracy:} Can the framework demonstrate improved benchmark performance on open-ended, deterministic mathematical reasoning tasks for models with no prior native Southeast Asian CoT capabilities, conditioned on native language alignment?
    \item \textbf{Efficiency of Training:} Does the framework provide improved training efficiency in terms of its required compute or dataset size? How does it compare to conventional supervised fine-tuning approaches that optimize multilingual performance on a massive SEA corpora?
    \item \textbf{Robustness and Generalizability:} Is the framework agnostic to model size and model family? Furthermore, does the acquisition of native multilingual CoT reasoning capabilities introduce catastrophic forgetting in high-resource base languages (e.g., English)?
\end{enumerate}

\subsection{Experimental Setup}
We evaluate our proposed approach using the base reasoning models \texttt{SmolLM3-3B}, \texttt{Qwen3-4B-Thinking-2507}, and \texttt{Qwen3-VL-8B-Thinking}.
For our training dataset, we extract a lean subset of 70,000 question-only samples from the \texttt{OpenMathReasoning-tir} corpus, discarding the accompanied reasoning and answer traces.
Each question is then paired with a user prompt presented natively in one of 7 target languages (\texttt{ZH,EN,Fi,IN,TA,TH,VI}), 
explicitly instructing the model to generate its intermediate CoT in that matching language.

To translate dynamically the long-context reasoning traces into specific target languages during training, 
we serve a local instance of \texttt{Gemma-SEA-LION-v4-27B-IT}, 
coupled with an agentic system $\mathcal{T}$ to optimize throughput and consistency (Figure~\ref{agentic_translator}).
This offers a functional alternative to rejection sampling on dynamic datasets. 
The temperature for translation initializes at $0.0$, incrementing by $0.1$ with each failed attempt (capped at $0.4$) for up to $10$ tries. 
Temperatures for scoring and language classification were fixed at $0.0$ to maintain deterministic evaluations.

We compare our post-trained models against the base models \texttt{SmolLM3-3B}, \texttt{Qwen3-4B-Thinking-2507}, and \texttt{Qwen3-VL-8B-Thinking},
as well as a multilingual model \texttt{Qwen-SEA-LION-v4-8B-VL}
of comparable size supervised fine-tuned on 9 million samples across different SEA languages.
We report the \textit{pass@5} and \textit{mean@5} scores to proxy reasoning intelligence and consistency, 
respectively, with standard deviations where applicable. 

Post-training experiments were conducted on an Ubuntu 22.04.5 LTS platform comprising four NVIDIA H200 GPUs (141,GB HBM3e VRAM) interconnected via NVLink.
The software stack consists of PyTorch 2.9.0, DeepSpeed 0.18.4, HuggingFace Transformers 4.57.1, and vLLM 0.13.0.
Total GPU runtime accumulated to 384 hours for \texttt{SmolLM3-3B}, 440 hours for \texttt{Qwen3-4B-Thinking-2507}, and 625 hours for \texttt{Qwen3-VL-8B-Thinking},
each inclusive of model training alongside the dedicated translator server.

\subsection{Evaluation Metrics}
To evaluate the native reasoning capabilities of LLMs across low-resource Southeast Asian languages, 
we introduce a comprehensive evaluation framework using the \texttt{AIME25} and \texttt{HMMT25} mathematical reasoning benchmarks (\citealp{aime25}; \citealp{dekoninck2026matharena}),
each translated into 7 different languages (\texttt{ZH,EN,Fi,IN,TA,TH,VI})
to accomodate a lack of multilingual benchmarks comprising all target variants.
The framework comprises two scoring modes, 
to evaluate both a model's linguistic biases and instruction-following capabilities.
Both modes evaluate on the same exact generated completions:
\begin{itemize}
    \item \textbf{Any-CoT:} In this mode, the model is scored solely on the correctness of its final parsed answer, regardless if its reasoning CoT exhibits linguistic drift to English or non-target languages.
    \item \textbf{Target-CoT:} In this mode, the model is scored on the correctness of its final parsed answer, as well as its adherence to the target languages during CoT reasoning per user instructions.
\end{itemize}

To analyze linguistic bias,
we also define Linguistic Retention Index \textbf{($\text{LRI}$)} as the probability a model successfully maintains its logical derivation process natively using the target language, 
rather than collapsing into a high-resource fallback.
Given a transition matrix $M \in \mathbb{R}^{N \times N}$, 
where the diagonal element $M_{i,i}$ denotes the probability that a model with target prompt language $\ell_i$ maintains its reasoning natively within $\ell_i$, 
the $\text{LRI}$ across a set of target languages $\mathcal{G}$ is computed as:
\begin{equation}
\text{LRI} = \frac{100}{|\mathcal{G}|} \sum_{\ell_i \in \mathcal{G}} M_{i,i}
\end{equation}

Thereafter, a high LRI for a given $|\mathcal{G}|$ languages bounded between $\text{0}$ and $\text{100}$ (inclusive) reflects native reasoning capabilities robust from biased linguistic preferences, 
whereas a low LRI signals systemic fallback behaviour into non-target languages.

For linguistic alignment verification of intermediate reasoning steps, 
we implement LLM-as-a-judge using the model \texttt{Gemma-SEA-LION-v4-27B-IT}.
We also truncate the first 70\% and tail 10\% of sequence lengths,
leaving only a 20\% critical snippet of reasoning traces for classification.
This prevents false positives induced by data leakage, often caused by model regurgitation of the native prompts near the beginning and end of its reasoning completion.
If language-mixing or switching were detected, the classification defaults strictly to high-resource languages, prioritizing English.

\subsection{Ablation Studies}
We conduct ablation studies using the base reasoning model \texttt{Qwen3-VL-8B-Thinking},
along with 4 distinct training configurations to isolate the core contributions of individual components within the OSCD framework,
specifically the secondary cosine distance loss $\mathcal{L}_{\text{JEPA}}$ as well as the agentic translator system $\mathcal{T}$.
The $\mathcal{L}_{\text{CE}}$ configuration accounts for the non-agentic deterministic translation of fine-tuned reasoning traces as a baseline 
referencing \citealp{chen2024breakinglanguagebarriersmultilingual2} and \citealp{barua2026longchainofthoughtreasoninglanguages},
less rejection sampling due to the dynamic nature of data localization involved.
For efficiency of experimentation, we use a representative, downscaled variant of the primary training dataset.
We construct a balanced subset of 7,500 samples, interleaved and uniformly distributed across 3 target languages (\texttt{ZH,EN,IN}) 
to enable a streamlined analysis of performance trade-offs. 

Accordingly, the \text{$\mathcal{L}_{\text{CE+JEPA (Agentic)}}$} configuration outperforms its other ablation variants, 
yielding the highest Target-CoT and LRI scores across both benchmarks (Table~\ref{overall-ablation-results}). 
Without the agentic translator to suppress zero-shot translation artifacts and variance, 
we observe increased noise in its respective gradient norms that led to severe performance degradations (Figure~\ref{ablation_curves}b),
in addition to slower inference speeds by a factor of two.
Similarly, without a secondary loss to mitigate cross-linguistic representational drifts, 
models consistently suffer from the divergence of its cosine distance metric (Figure~\ref{ablation_curves}d),
causing a wider spread in reasoning accuracies and lower LRI scores despite a small set of 3 languages. 
This validates both our initial hypothesis and the robustness of our training framework.

\begin{table*}[t]
\centering
\small
\setlength{\tabcolsep}{3pt}
\begin{tabular*}{\textwidth}{@{\extracolsep{\fill}}l c c c c c c c c c c}
\hline
 & \multicolumn{5}{c}{\textbf{AIME25}} & \multicolumn{5}{c}{\textbf{HMMT25}} \\
\cline{2-6} \cline{7-11}
 & \multicolumn{2}{c}{\textbf{Any-CoT}} & \multicolumn{3}{c}{\textbf{Target-CoT}} & \multicolumn{2}{c}{\textbf{Any-CoT}} & \multicolumn{3}{c}{\textbf{Target-CoT}} \\
\cline{2-3} \cline{4-6} \cline{7-8} \cline{9-11}
\textbf{Model} & \textbf{Pass@5} & \textbf{Mean@5} & \textbf{Pass@5} & \textbf{Mean@5} & \textbf{LRI} & \textbf{Pass@5} & \textbf{Mean@5} & \textbf{Pass@5} & \textbf{Mean@5} & \textbf{LRI} \\
\hline
SmolLM3-3B & 41.9 {\scalebox{0.8}{$\pm$15.6}} & 23.2 {\scalebox{0.8}{$\pm$10.4}} & 12.9 {\scalebox{0.8}{$\pm$23.6}} & 6.8 {\scalebox{0.8}{$\pm$13.0}} & 27.4 & 22.9 {\scalebox{0.8}{$\pm$7.6}} & 12.0 {\scalebox{0.8}{$\pm$5.5}} & 7.1 {\scalebox{0.8}{$\pm$13.1}} & 3.8 {\scalebox{0.8}{$\pm$7.8}} & 27.3 \\
\cellcolor{white!15}\textbf{Ours-SmolLM3-3B} & 31.4 {\scalebox{0.8}{$\pm$14.5}} & 18.6 {\scalebox{0.8}{$\pm$8.4}} & 29.5 {\scalebox{0.8}{$\pm$14.8}} & 14.4 {\scalebox{0.8}{$\pm$8.8}} & \textbf{\color{red}89.8} & 13.3 {\scalebox{0.8}{$\pm$9.2}} & 7.0 {\scalebox{0.8}{$\pm$5.5}} & 12.4 {\scalebox{0.8}{$\pm$9.2}} & 6.1 {\scalebox{0.8}{$\pm$4.7}} & \textbf{\color{red}91.8} \\
Qwen3-4B-Thinking-2507 & 83.8 {\scalebox{0.8}{$\pm$2.3}} & 69.0 {\scalebox{0.8}{$\pm$8.0}} & 22.9 {\scalebox{0.8}{$\pm$39.2}} & 18.9 {\scalebox{0.8}{$\pm$33.1}} & 26.9 & 52.9 {\scalebox{0.8}{$\pm$5.2}} & 41.0 {\scalebox{0.8}{$\pm$5.0}} & 15.2 {\scalebox{0.8}{$\pm$26.1}} & 10.9 {\scalebox{0.8}{$\pm$19.1}} & 25.8 \\
\cellcolor{white!15}\textbf{Ours-Qwen3-4B} & \textbf{\color{red}85.2 {\scalebox{0.8}{$\pm$2.6}}} & \textbf{\color{red}69.0 {\scalebox{0.8}{$\pm$6.2}}} & \textbf{\color{red}68.6 {\scalebox{0.8}{$\pm$13.7}}} & 42.5 {\scalebox{0.8}{$\pm$16.7}} & 59.7 & \textbf{\color{red}54.8 {\scalebox{0.8}{$\pm$2.6}}} & 41.7 {\scalebox{0.8}{$\pm$4.3}} & 41.0 {\scalebox{0.8}{$\pm$13.8}} & 23.4 {\scalebox{0.8}{$\pm$12.9}} & 55.5 \\
Qwen-SEA-LION-v4-8B-VL & 49.5 {\scalebox{0.8}{$\pm$10.1}} & 32.2 {\scalebox{0.8}{$\pm$8.1}} & 38.6 {\scalebox{0.8}{$\pm$14.6}} & 21.0 {\scalebox{0.8}{$\pm$11.2}} & 65.8 & 31.0 {\scalebox{0.8}{$\pm$10.7}} & 17.4 {\scalebox{0.8}{$\pm$8.1}} & 22.9 {\scalebox{0.8}{$\pm$12.5}} & 10.6 {\scalebox{0.8}{$\pm$8.0}} & 65.5 \\
Qwen3-VL-8B-Thinking & 83.3 {\scalebox{0.8}{$\pm$4.7}} & 68.6 {\scalebox{0.8}{$\pm$5.8}} & 21.4 {\scalebox{0.8}{$\pm$37.2}} & 17.8 {\scalebox{0.8}{$\pm$31.2}} & 24.0 & 52.4 {\scalebox{0.8}{$\pm$5.3}} & \textbf{\color{red}42.4 {\scalebox{0.8}{$\pm$4.2}}} & 13.8 {\scalebox{0.8}{$\pm$24.1}} & 11.1 {\scalebox{0.8}{$\pm$20.1}} & 23.8 \\
\cellcolor{white!15}\textbf{Ours-Qwen3-VL-8B} & 73.8 {\scalebox{0.8}{$\pm$11.0}} & 60.9 {\scalebox{0.8}{$\pm$13.6}} & \textbf{\color{red}68.6 {\scalebox{0.8}{$\pm$14.0}}} & \textbf{\color{red}47.5 {\scalebox{0.8}{$\pm$17.9}}} & 86.0 & 43.8 {\scalebox{0.8}{$\pm$12.2}} & 33.2 {\scalebox{0.8}{$\pm$10.5}} & \textbf{\color{red}41.4 {\scalebox{0.8}{$\pm$13.7}}} & \textbf{\color{red}25.6 {\scalebox{0.8}{$\pm$11.7}}} & 84.9 \\
\hline
\end{tabular*}
\caption{\label{overall-benchmark-results}
Overall main performance comparison across \texttt{AIME25}, \texttt{HMMT25} benchmarks evaluating Any-CoT vs Target-CoT setups, 
averaged across 7 target languages (\texttt{ZH,EN,Fi,IN,TA,TH,VI}) 
comparing base reasoning models and OSCD post-trained models. 
}
\end{table*}

\begin{figure*}[t]  
    \centering
    \includegraphics[width=0.95\textwidth]{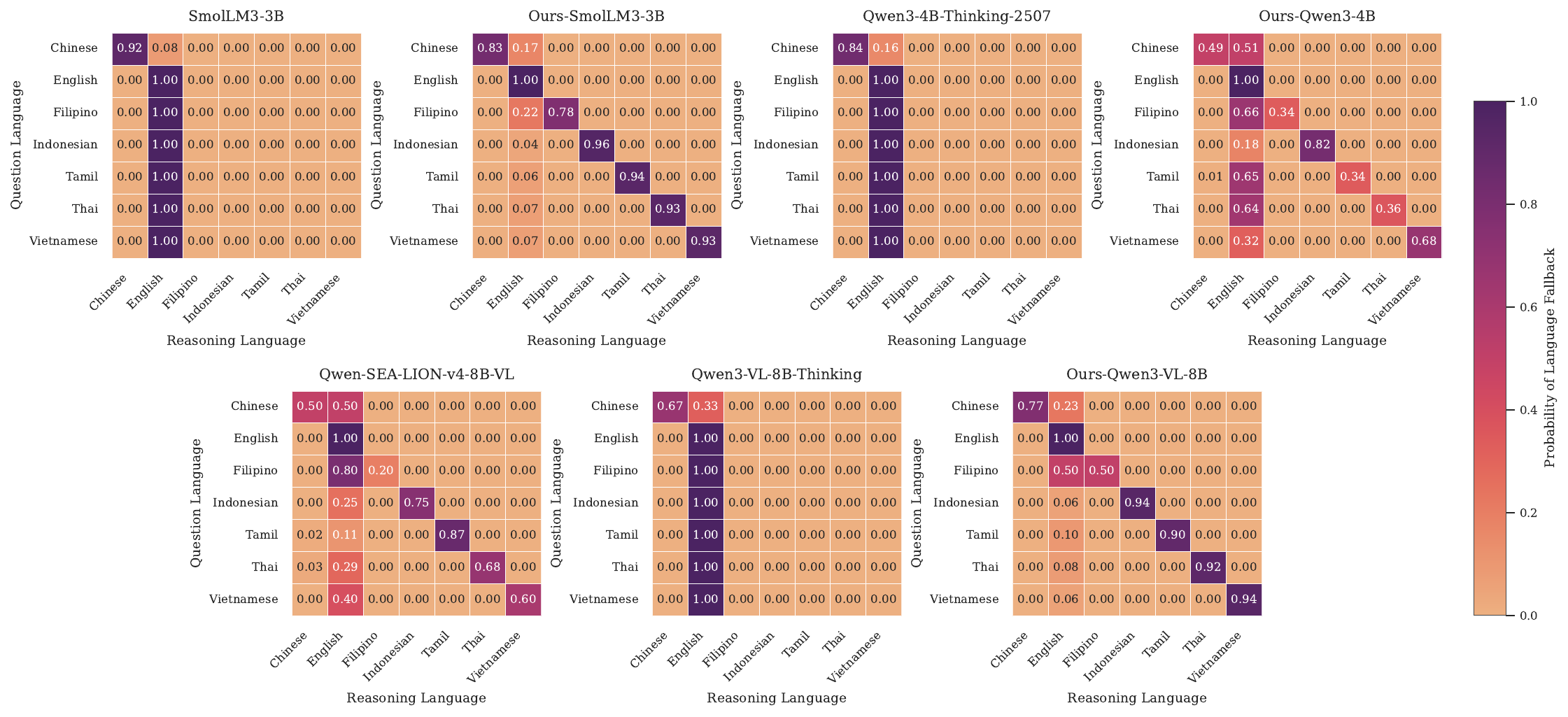} 
    \captionsetup{justification=centering}
    \caption{Overall main language fallback matrices 
    comparing base reasoning models and OSCD post-trained models across \texttt{AIME25}, \texttt{HMMT25} benchmarks and 7 target languages (\texttt{ZH,EN,Fi,IN,TA,TH,VI}). 
    }
    \label{main_linguistic_alignment_matrices_overall}
\end{figure*}

\begin{figure*}[t]  
    \centering
    \includegraphics[width=0.9\textwidth]{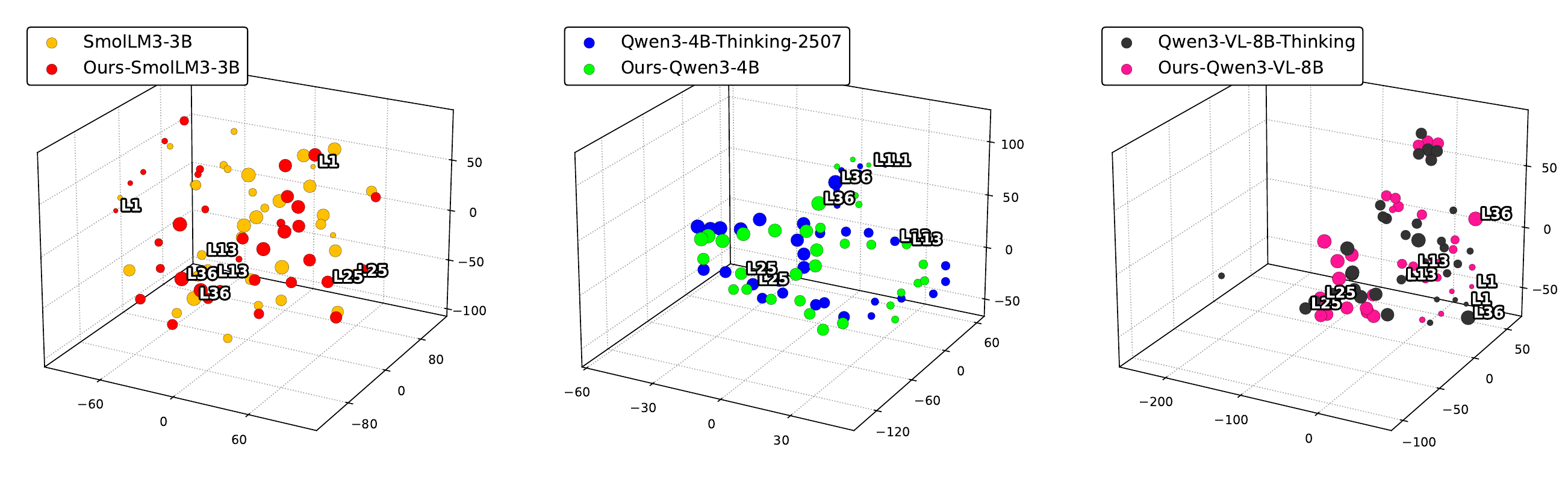} 
    \captionsetup{justification=centering}
    \caption{3D t-SNE plots of layer-wise hidden activations comparing base reasoning models and OSCD post-trained models, 
    using the \texttt{AIME25} dataset across 7 target languages (\texttt{ZH,EN,Fi,IN,TA,TH,VI}) as input corpus.
    The layers $1$ to $N$ are indicated by expanding marker sizes.
    }
    \label{layer_trajectories_side_by_side_aime25}
\end{figure*}

\begin{figure*}[t]  
    \centering
    \includegraphics[width=0.9\textwidth]{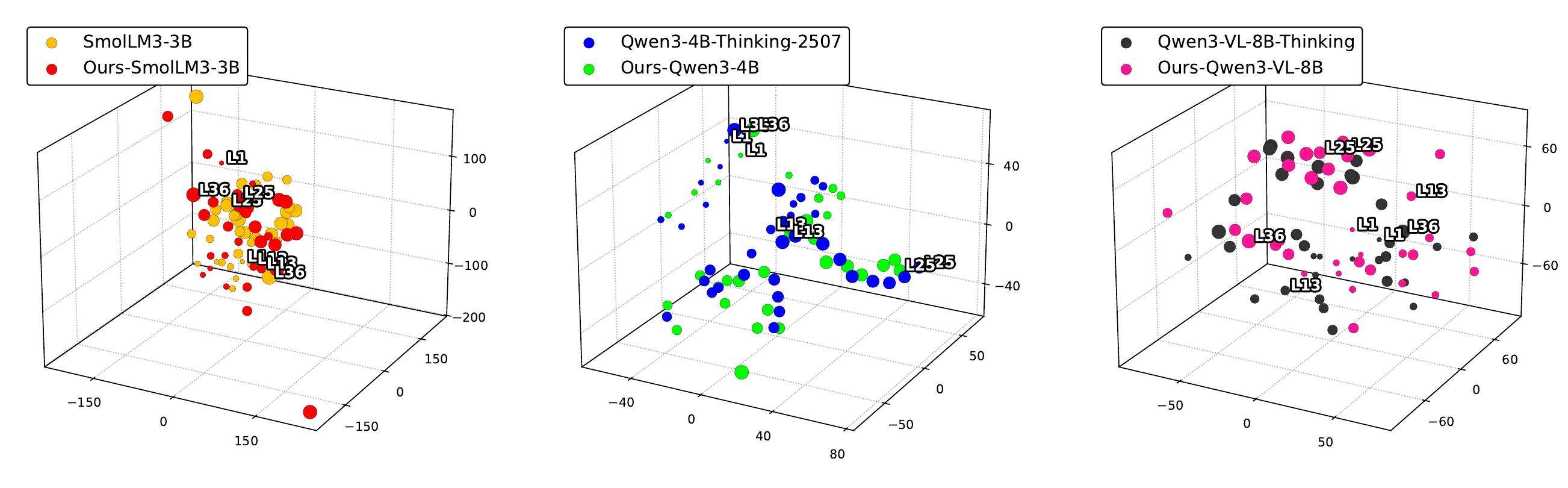} 
    \captionsetup{justification=centering}
    \caption{3D t-SNE plots of layer-wise hidden activations comparing base reasoning models and OSCD post-trained models, 
    using the \texttt{HMMT25} dataset across 7 target languages (\texttt{ZH,EN,Fi,IN,TA,TH,VI}) as input corpus.
    The layers $1$ to $N$ are indicated by expanding marker sizes.
    }
    \label{layer_trajectories_side_by_side_hmmt25}
\end{figure*}

\subsection{Low-Resource Native Reasoning}
The robustness of model native reasoning performance in low-resource SEA languages improves substantially with OSCD post-training.
As shown in Table~\ref{overall-benchmark-results}, 
we observe an average 2--3 times overall improvement in both \textit{Pass@5} and \textit{Mean@5} Target-CoT scores
across both \texttt{AIME25} and \texttt{HMMT25} mathematical benchmarks,
reflecting the successful cross-lingual transfer of high-resource base capabilities into low-resource target languages (Table~\ref{overall-benchmark-results}).
Importantly, the baseline reasoning models exhibit an English sinking state as default fallback language, 
with a minor exception for Chinese language due to abundance in its pre-training web corpus (Figure~\ref{main_linguistic_alignment_matrices_overall}).
Our models, however, demonstrate consistent mitigation of this bias, 
achieving a 2.2--3.6$\times$ improvement in LRI relative to model baselines.

With Any-CoT, a mild reduction in overall scores for the 
\texttt{Ours-SmolLM3-3B} and \texttt{Ours-Qwen3-VL-8B} models were observed,
attributed to the alteration of its natural fallback language.
Isolated for high-resource languages, however,
the performance drops remain marginal compared to baseline (Figure~\ref{bar_plots}).
This suggests that foundational reasoning capabilities remain well-preserved, 
with noticeable Any-CoT reductions largely driven by a linguistic shift toward a more unbiased, 
multilingual distribution.
In contrast, \texttt{Ours-Qwen3-4B} demonstrated modest improvements under Any-CoT settings,
accompanied by a 40\% probability of non-target fallback behavior;
indicating a continued tendency to optimize reasoning trajectories for accuracy via English,
while concurrently benefiting from knowledge expansion across low-resource settings.

\subsection{Training Efficiency and Generalizability}
A primary advantage of the OSCD training framework is sample efficiency, 
which leverages on stable and dynamically localized synthetic completions to mitigate the scarcity of quality data in low-resource target languages.
To evaluate this, we compare our models against \texttt{Qwen-SEA-LION-v4-8B-VL}, 
a competitive baseline trained on 9 million high-quality instruction-text pairs spanning English and SEA languages.
On the other hand, our models utilize a significantly smaller dataset of only 70,000 samples, 
distributed uniformly across 7 languages.
Despite a raw dataset size reduction of over 90\%, our framework yields substantial improvements in LRI metrics (Table~\ref{overall-benchmark-results}).
\texttt{Ours-Qwen3-VL-8B}, in particular, surpasses \texttt{Qwen-SEA-LION-v4-8B-VL} in LRI by 30\%,
thus demonstrating the robustness of dynamic OSCD over static SFT pipelines for multilingual acquisition.

To evaluate the generalizability of OSCD, 
we experiment across three parameter scales (i.e., 3B, 4B, and 8B),
spanning two distinct model families (i.e., Qwen3 and SmolLM3) as well as both text-only and multimodal dense architectures.
Although our compute constraints prevented scaling beyond 8B parameters, 
we observed consistent improvements across different models. 
Specifically, an average 2--3 times overall improvement in Target-CoT reasoning performance across different benchmarks,
as well as an increase in LRI up to 3.6 times corresponding to the
successfully neutralization of high-resource English bias present in baseline models.
This provides a working foundation for subsequent post-training,
particularly with open-ended downstream applications that involve regional and cultural contexts.

\subsection{Scaling Behaviour and Hidden Activations}
Our scaling analysis reveals a non-linear behavior, 
driven by long-context reasoning capabilities relative to model weight sizes.
\texttt{Ours-SmolLM3-3B}, for instance, lacks the capacity for extended generations 
which limits its exposure duration to non-target linguistic fallbacks.
A substantially higher LRI is hence achieved, though its Target-CoT improvements remain modest. 
\texttt{Ours-Qwen3-4B}, on the other hand, features extended reasoning capabilities but remains bottlenecked by its compact size,
therefore compromising on LRI to sustain the accuracy of its answers.
Among all, \texttt{Ours-Qwen3-VL-8B} provides the optimal foundation,
supporting the multilingual expansion of knowledge space while simultaneously 
resisting non-target linguistic fallbacks over prolonged generations.

Figures~\ref{layer_trajectories_side_by_side_aime25} and~\ref{layer_trajectories_side_by_side_hmmt25} 
illustrate the layer-wise hidden-state activations comparing base reasoning models and OSCD post-trained models, 
using the \texttt{AIME25} and \texttt{HMMT25} datasets across 7 target languages as input corpus, respectively. 
In general, it is observed the initial layers remain clustered and well-aligned between models before and after post-training, 
while geometric shifts occur predominantly within the middle-to-later layers.
The observation reflects a greater dependency on generative steering along deeper layers for downstream benchmark improvements, 
as opposed to multilingual comprehension across problems in low-resource languages,
particularly with baseline models of larger sizes and stronger default capabilities.
With smaller models, however, substantial shifts can emerge along the early representational layers, 
indicating a fundamental alteration to how challenging problems are semantically comprehended as new knowledge gets ingested.

\section{Conclusion}
To conclude, we present the OSCD algorithm which enables 2--3 times substantial improvements in Target-CoT reasoning performance,
conditioned on linguistic alignment verifications across low-resource Southeast Asian languages. 
The enabling of native multilingual reasoning capabilities proved largely additive,
with limited observations of catastrophic forgetting observed in high-resource base capabilities. 
Crucially, we extend the upper-bound intelligence potential for native CoT reasoning in low-resource languages, 
offering the open community an additional layer of training foundation for subsequent reinforcement learning. 
This empowers non-native English speakers the ability to leverage advanced AI capabilities, 
thereby improving system usability and broad accessibility. 
Future research will focus on integrating reinforcement learning and cultural fine-tuning for downstream applications beyond deterministic reasoning. 
This includes scaling the approach to open-ended tasks that heavily rely on regional and cultural contexts.

\section*{Acknowledgments}
This work was supported by the DSO-AISG Incentive Award. 
The views expressed are solely those of the authors and do not represent the opinions of DSO National Laboratories or AI Singapore.

\bibliography{aaai2027}

\clearpage 
\appendix

\section{Training Details}
\label{training}

\subsection{Algorithm Pseudocode}
The details of training process are provided in Algorithm~\ref{cot_mutation_distill}.

\subsection{Hyperparameters}
Table~\ref{hyperparams} provides the hyperparameter settings for post-training experiments. 
Inference parameters for generative rollouts and reasoning benchmarks were consistent with the default recommendations of the respective model developers.

\begin{table}[H]
\centering
\small
\captionsetup{justification=raggedright, singlelinecheck=false}
\begin{tabular}{lc} 
\toprule
\textbf{Hyperparameter} & \textbf{Value} \\
\midrule
\textit{Optimization:} & \\
Optimizer & AdamW \\
Learning rate & $2 \times 10^{-5}$ \\
\midrule
\textit{Training:} & \\
Training epochs & 1 \\
Precision & BF16 \\
Gradient clip norm & 1.0 \\
\midrule
\textit{Learning Rate Schedule:} & \\
Scheduler type & Cosine \\
Warmup fraction & 0.1 \\
\midrule
\textit{Loss Weights:} & \\
Cross-Entropy ($\lambda_{CE}$) & 1.0 \\
JEPA ($\lambda_{JEPA}$) & 1.0 \\
\midrule
\textit{Sequence Lengths:} & \\
Max prompt tokens \textit{(Training)} & 4096 \\
Max completion tokens \textit{(Training)} & 8192 \\
Max completion tokens \textit{(Benchmark)} & 81920 \\
\bottomrule
\end{tabular}
\caption{Hyperparameter settings.}
\label{hyperparams}
\end{table}

\subsection{Compute Resources}
\begin{itemize}
  \item \textbf{Hardware:} $4 \times$ NVIDIA H200 GPUs (141\,GB HBM3e VRAM each) interconnected via NVLink.
  \item \textbf{Software:} PyTorch 2.9.0, DeepSpeed 0.18.4, HuggingFace Transformers 4.57.1, and vLLM 0.13.0.
  \item \textbf{GPU Runtime:} 384 hours for \texttt{SmolLM3-3B}, 440 hours for \texttt{Qwen3-4B-Thinking-2507}, and 625 hours for \texttt{Qwen3-VL-8B-Thinking}. \\
  (\textit{Note: Total runtime accounts for both the target model training and local vLLM server for dynamic translations.})
\end{itemize}

\subsection{Prompt Templates}
\label{prompts}

This appendix provides example prompt templates and scaffolds used within this study. 
Specifically, Table~\ref{classification_prompt} provides the language classification prompt, 
Table~\ref{translation_prompt} the zero-shot translation template, 
and Table~\ref{evaluation_prompt} the quality evaluation framework.
All scaffolds were integrated into our agentic translator during training. 
For benchmarks involving language classification, 
we rely exclusively on Table~\ref{classification_prompt}. 

\section{Supplementary Results}
\label{supplementary}

\subsection{Detailed Benchmarks}

\begin{figure*}[t] 
    \centering
    \includegraphics[width=\textwidth]{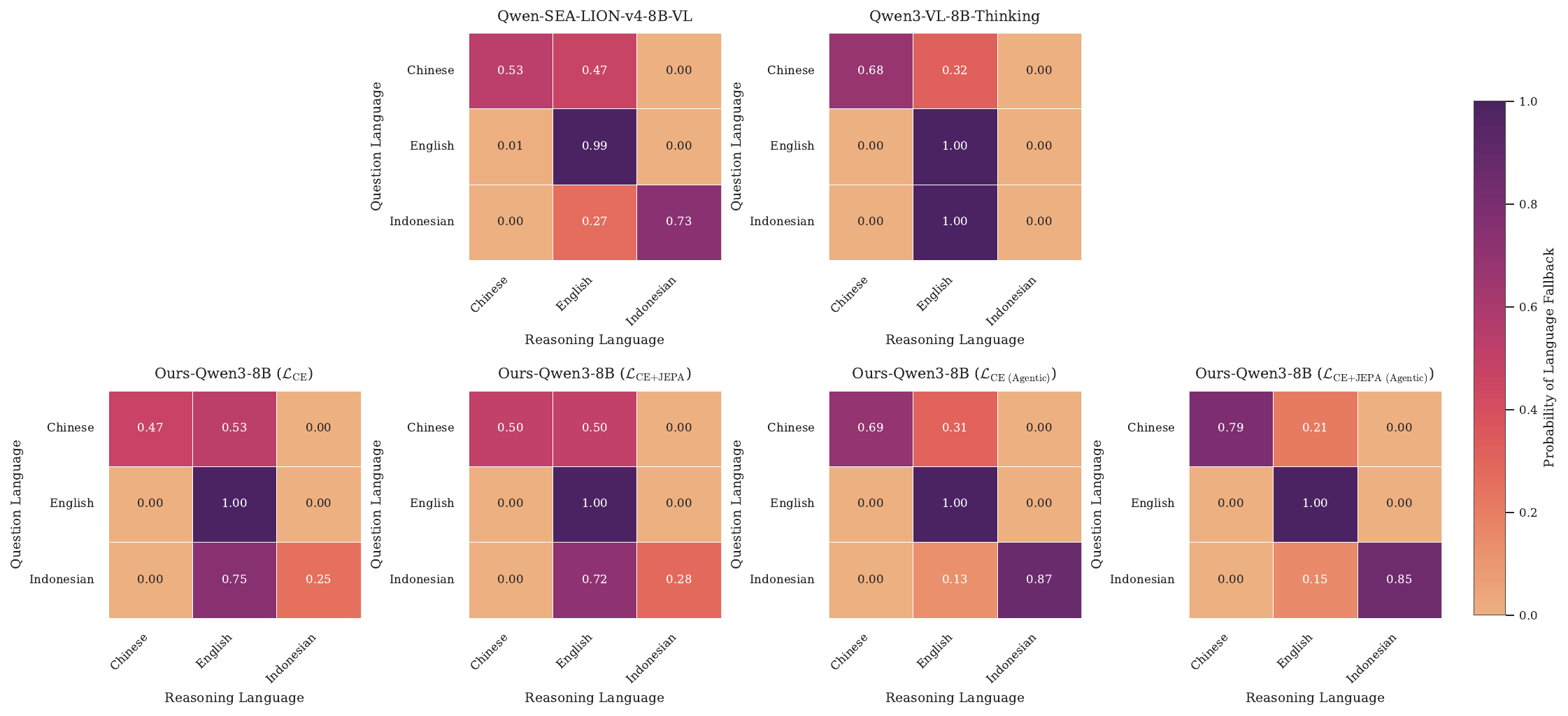} 
    \captionsetup{justification=centering}
    \caption{Ablation language fallback matrices on the \texttt{AIME25} benchmark 
    comparing different training loss configurations across 3 target languages (\texttt{ZH,EN,IN}). 
    }
    \label{ablation_aime25_matrix}
\end{figure*}

\begin{figure*}[t]  
    \centering
    \includegraphics[width=\textwidth]{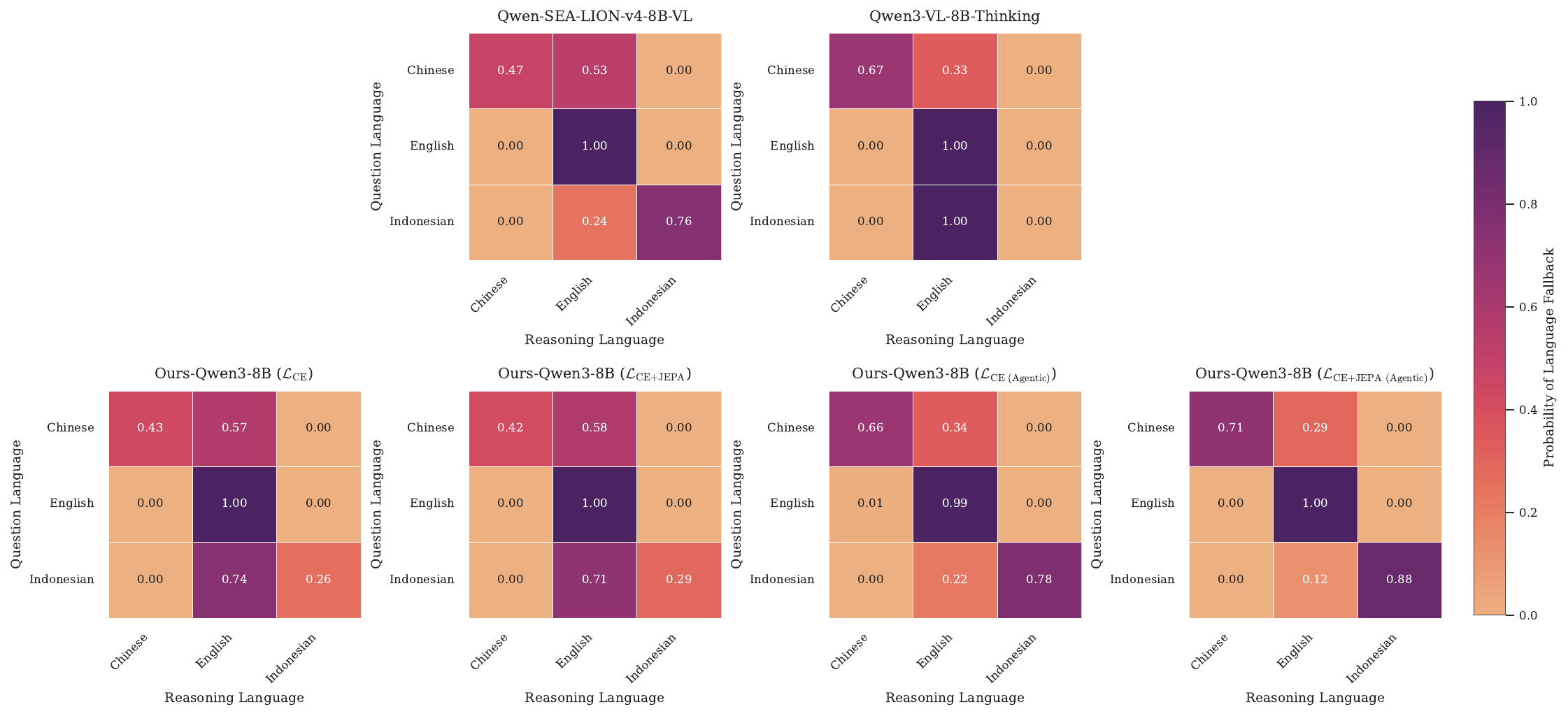} 
    \captionsetup{justification=centering}
    \caption{Ablation language fallback matrices on the \texttt{HMMT25} benchmark 
    comparing different training loss configurations across 3 target languages (\texttt{ZH,EN,IN}). 
    }
    \label{ablation_hmmt25_matrix}
\end{figure*}

\begin{figure*}[t]  
    \centering
    \includegraphics[width=\textwidth]{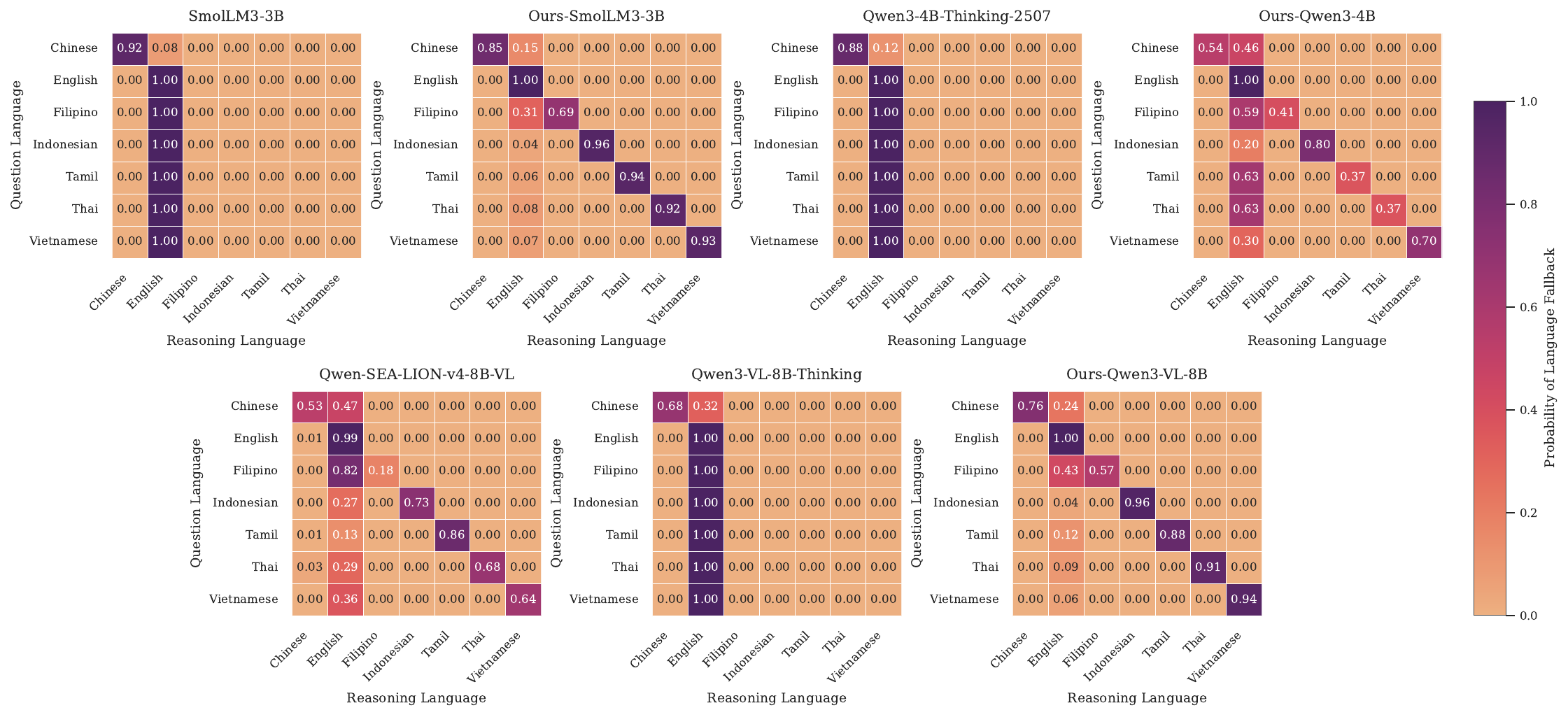} 
    \captionsetup{justification=centering}
    \caption{Main language fallback matrices 
    comparing base reasoning models and OSCD post-trained models across the \texttt{AIME25} benchmark
    and 7 target languages (\texttt{ZH,EN,Fi,IN,TA,TH,VI}). 
    }
    \label{main_linguistic_alignment_matrices_aime25}
\end{figure*}

\begin{figure*}[t] 
    \centering
    \includegraphics[width=\textwidth]{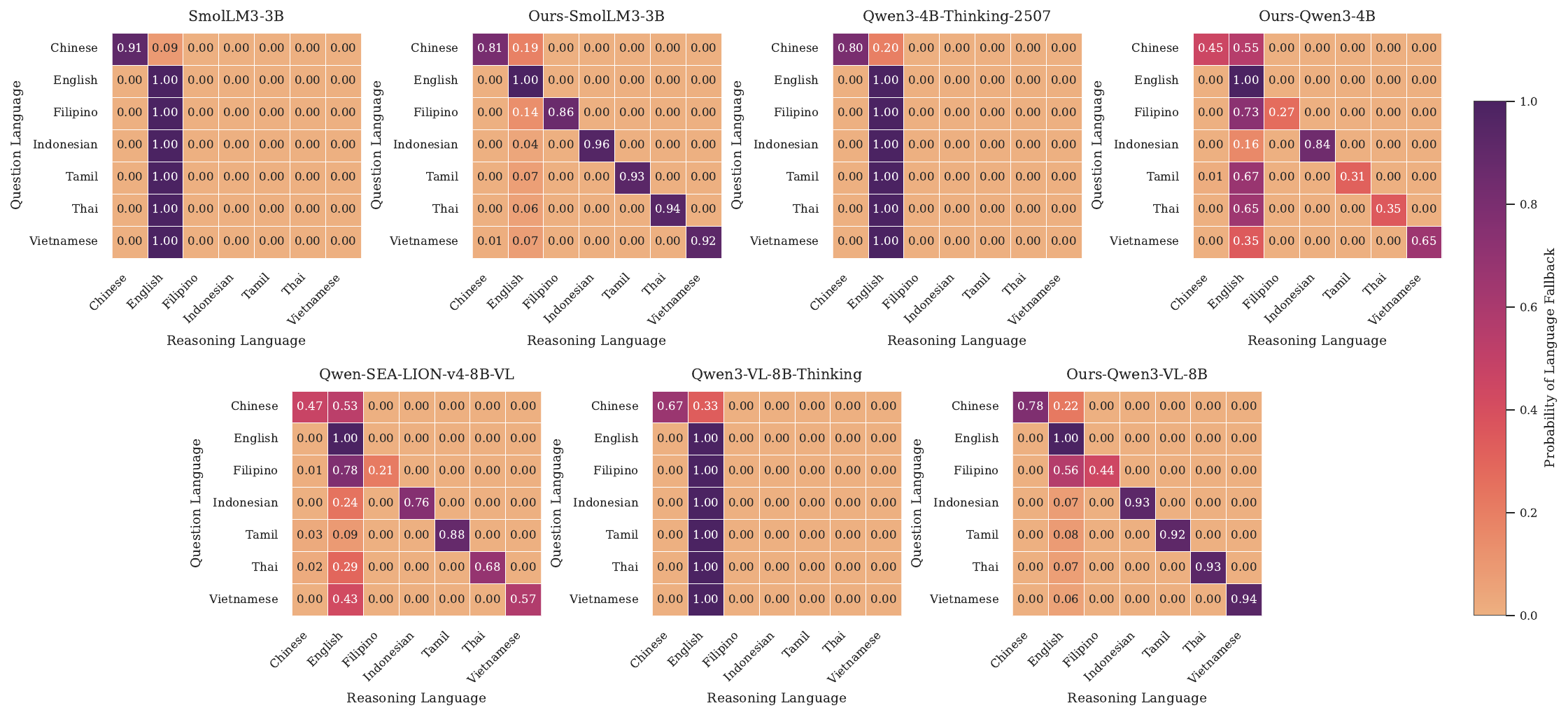} 
    \captionsetup{justification=centering}
    \caption{Main language fallback matrices 
    comparing base reasoning models and OSCD post-trained models across the \texttt{HMMT25} benchmark
    and 7 target languages (\texttt{ZH,EN,Fi,IN,TA,TH,VI}). 
    }    
    \label{main_linguistic_alignment_matrices_hmmt25}
\end{figure*}

This appendix provides benchmark results for both our ablation studies and main experiments.
Tables~\ref{aime25-ablation-micro}--\ref{hmmt25-ablation-micro} provide the ablation results for \texttt{AIME25} and \texttt{HMMT25}, respectively,
with Figures~\ref{ablation_aime25_matrix}--\ref{ablation_hmmt25_matrix} illustrating language fallback matrices comparing different model training configurations across the 3 target languages (\texttt{ZH,EN,IN}).
Tables~\ref{aime25-main-any}--\ref{hmmt25-main-target} provide the main results for \texttt{AIME25} and \texttt{HMMT25}, respectively,
with Figures~\ref{main_linguistic_alignment_matrices_aime25}--\ref{main_linguistic_alignment_matrices_hmmt25} 
illustrating language fallback matrices comparing different model training configurations across the 7 target languages (\texttt{ZH,EN,Fi,IN,TA,TH,VI}).

\subsection{Training Dynamics}
\label{train_dynamics}
\begin{figure*}[t] 
    \centering
    
    \begin{subfigure}{\textwidth}
        \centering
        \includegraphics[width=0.955\textwidth]{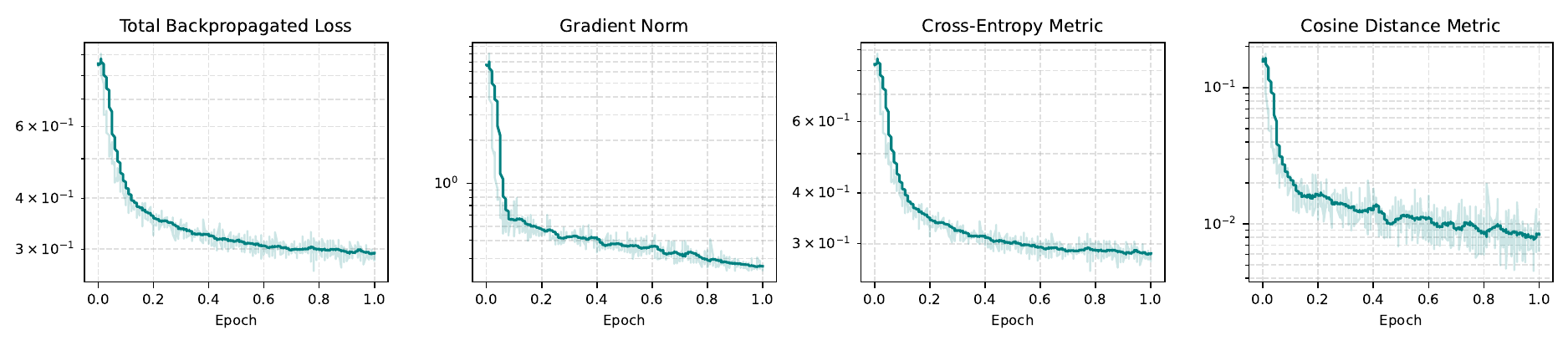}
        \caption{\texttt{Ours-SmolLM3-3B}}
        \label{model_x}
    \end{subfigure}
    \vspace{0.01cm} 

    \begin{subfigure}{\textwidth}
        \centering
        \includegraphics[width=0.95\textwidth]{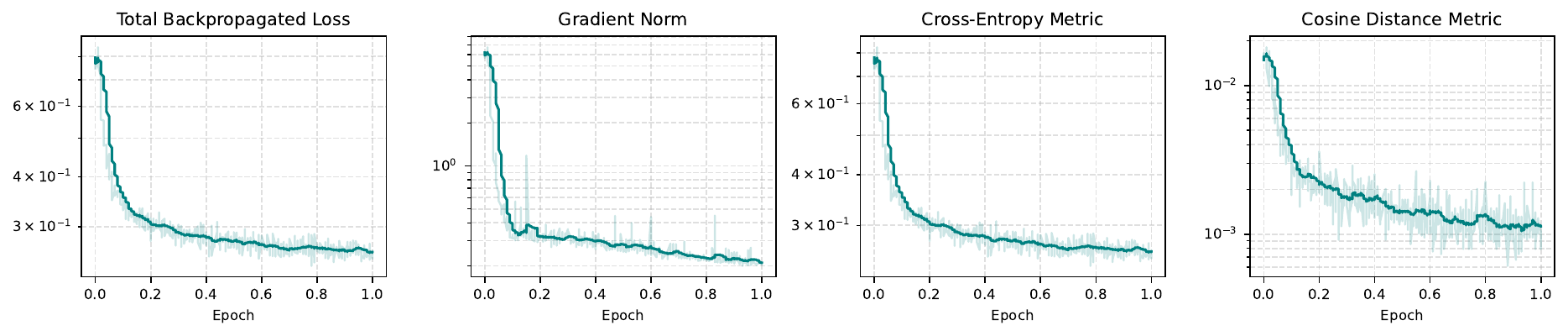}
        \caption{\texttt{Ours-Qwen3-4B}}
        \label{model_y}
    \end{subfigure}
    \vspace{0.01cm}

    \begin{subfigure}{\textwidth}
        \centering
        \includegraphics[width=0.95\textwidth]{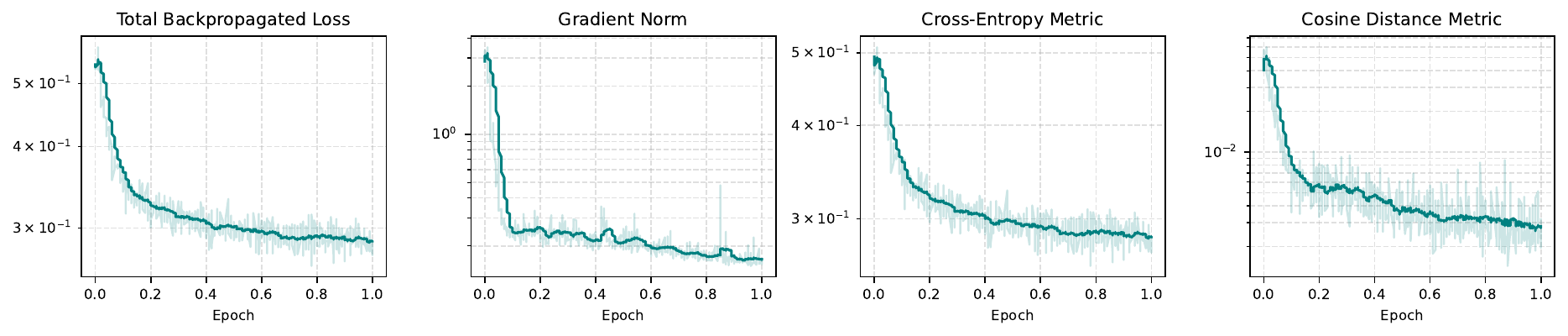}
        \caption{\texttt{Ours-Qwen3-VL-8B}}
        \label{model_z}
    \end{subfigure}
    \vspace{0.01cm}

    \captionsetup{justification=centering}
    \caption{Training dynamics for models (a) \texttt{SmolLM3-3B}, (b) \texttt{Qwen3-4B-Thinking-2507}, and (c) \texttt{Qwen3-VL-8B-Thinking} using the OSCD post-training framework.}
    \label{qwen8b_train_metrics}
\end{figure*}

Figure~\ref{qwen8b_train_metrics} plots the total backpropagated loss, gradient norm, cross-entropy, and cosine distance metrics for our main training experiments.

\subsection{Hidden Representations}
\label{hidden_rep}
\begin{figure*}[t] 
    \centering
    
    \begin{subfigure}{\textwidth}
        \centering
        \includegraphics[width=0.8\textwidth]{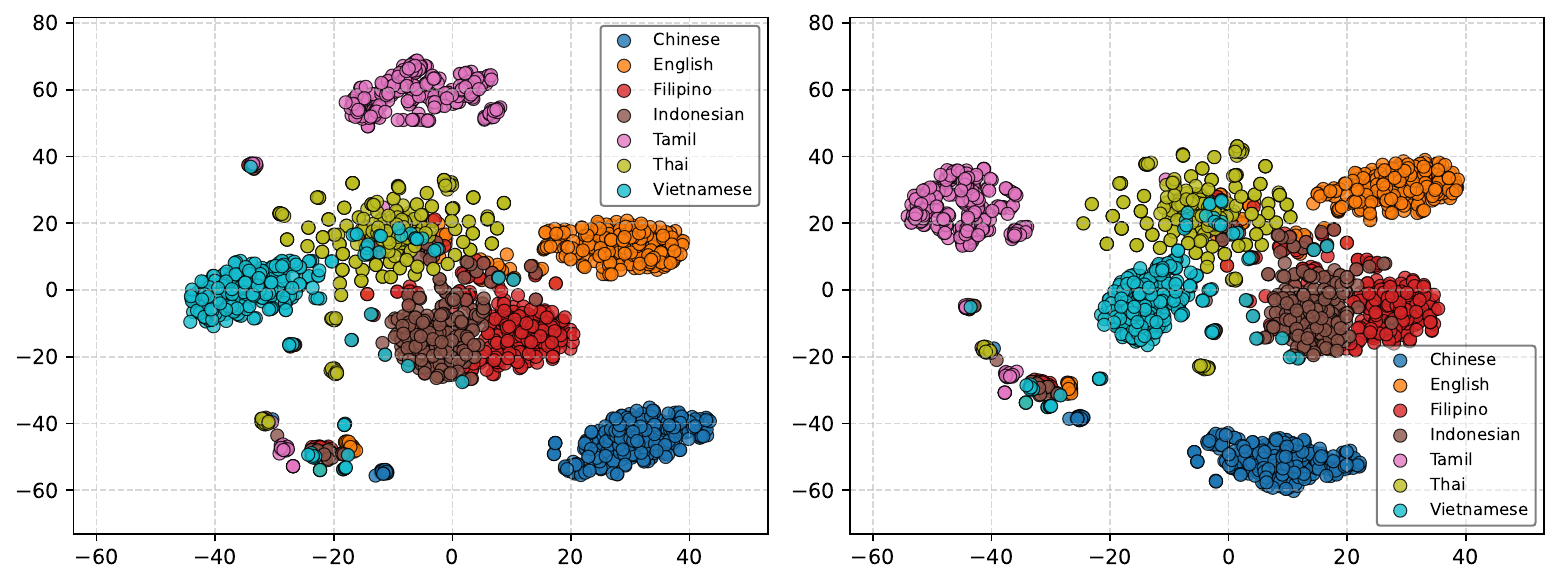}
        \caption{\texttt{SmolLM3-3B} (\textit{left}) and \texttt{Ours-SmolLM3-3B} (\textit{right})}
        \label{tsne2D_model_x_L1}
    \end{subfigure}
    \vspace{0.01cm} 

    \begin{subfigure}{\textwidth}
        \centering
        \includegraphics[width=0.8\textwidth]{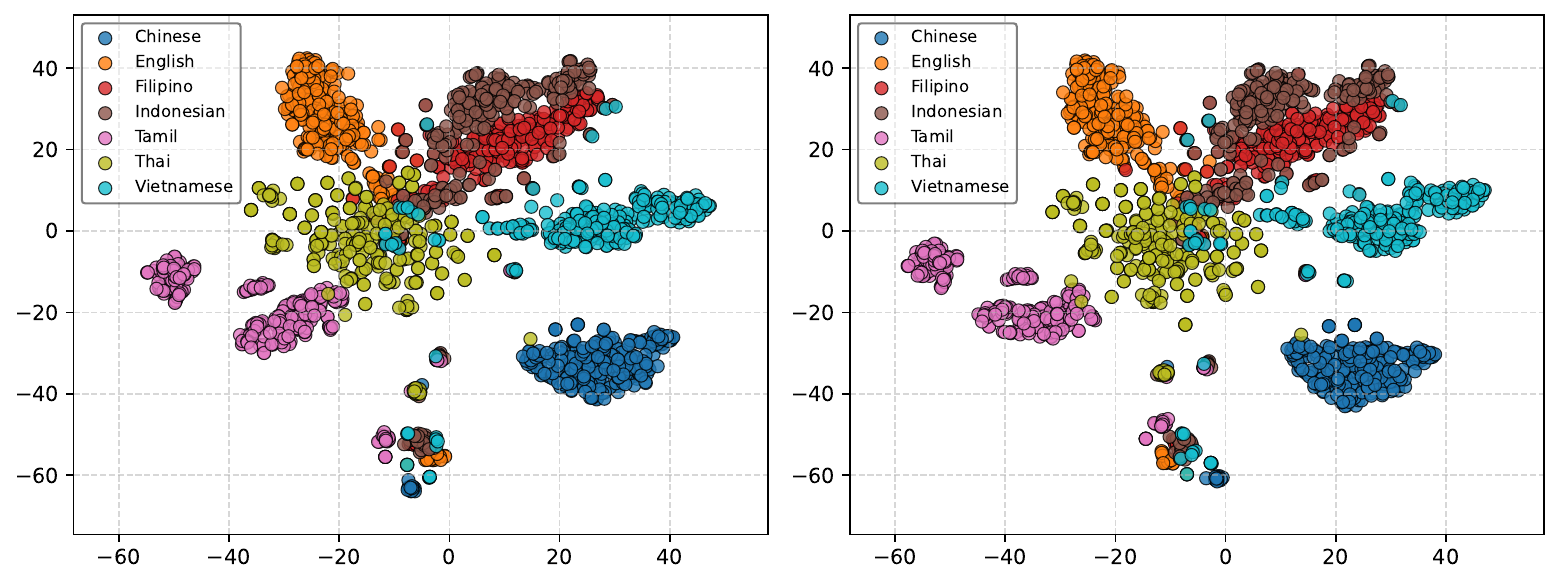}
        \caption{\texttt{Qwen3-4B-Thinking-2507} (\textit{left}) and \texttt{Ours-Qwen3-4B} (\textit{right})}
        \label{tsne2D_model_y_L1}
    \end{subfigure}
    \vspace{0.01cm}

    \begin{subfigure}{\textwidth}
        \centering
        \includegraphics[width=0.8\textwidth]{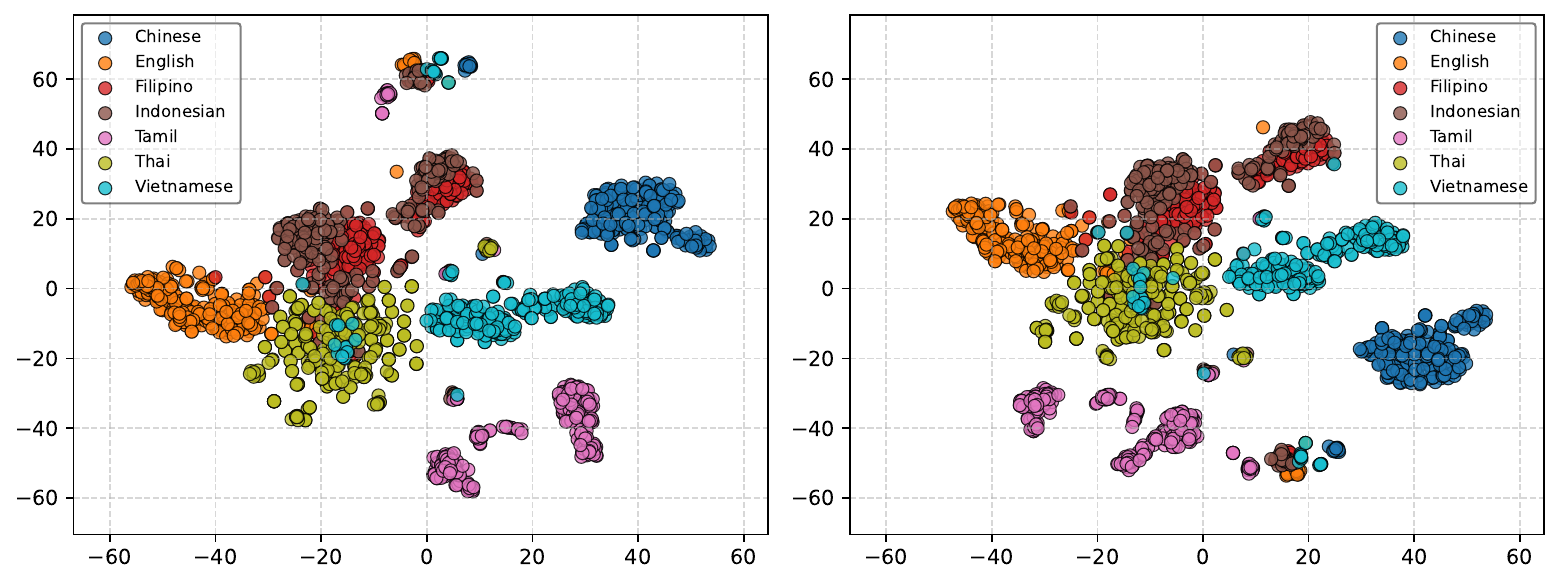}
        \caption{\texttt{Qwen3-VL-8B-Thinking} (\textit{left}) and \texttt{Ours-Qwen3-VL-8B} (\textit{right})}
        \label{tsne2D_model_z_L1}
    \end{subfigure}
    \vspace{0.01cm}

    \captionsetup{justification=centering}
    \caption{2D t-SNE visualizations of the first layer (\textit{Layer 1}) multilingual representations for (a) \texttt{SmolLM3-3B}, (b) \texttt{Qwen3-4B-Thinking-2507}, and (c) \texttt{Qwen3-VL-8B-Thinking} before (\textit{left}) and after (\textit{right}) post-training with OSCD.}
    \label{tsne2D_L1}
\end{figure*}

\begin{figure*}[t] 
    \centering
    
    \begin{subfigure}{\textwidth}
        \centering
        \includegraphics[width=0.8\textwidth]{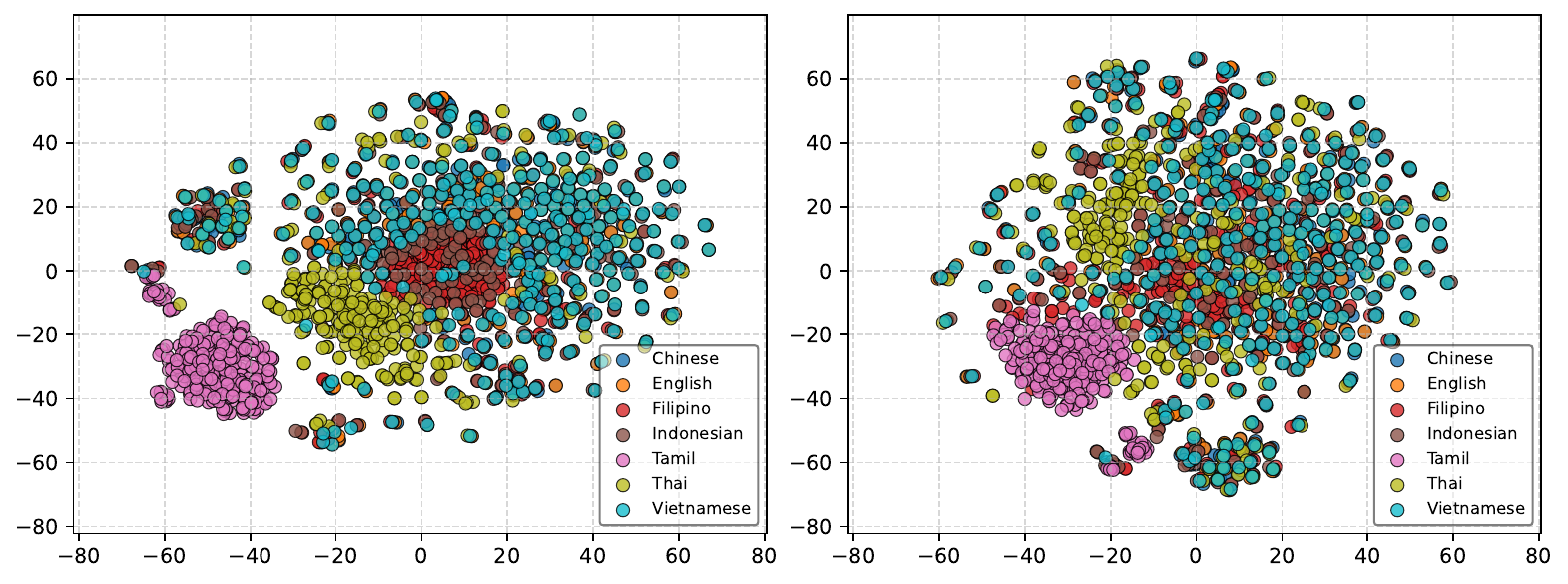}
        \caption{\texttt{SmolLM3-3B} (\textit{left}) and \texttt{Ours-SmolLM3-3B} (\textit{right})}
        \label{tsne2D_model_x_L13}
    \end{subfigure}
    \vspace{0.01cm} 

    \begin{subfigure}{\textwidth}
        \centering
        \includegraphics[width=0.8\textwidth]{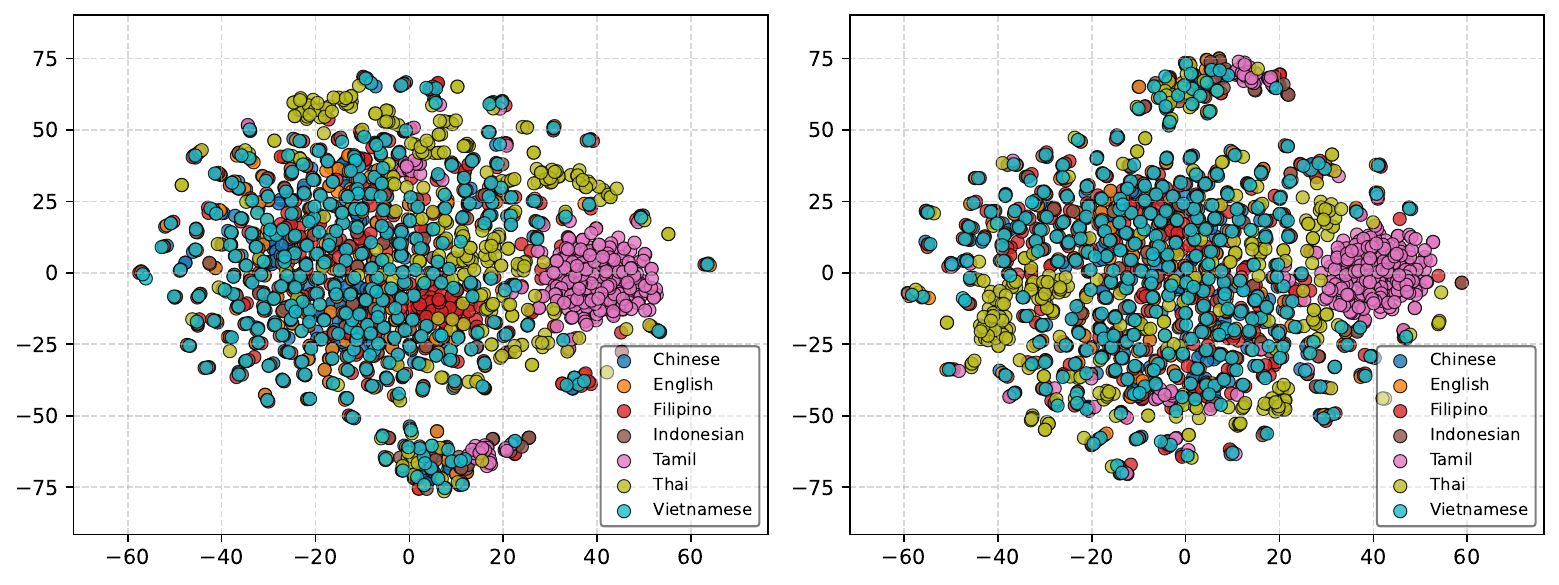}
        \caption{\texttt{Qwen3-4B-Thinking-2507} (\textit{left}) and \texttt{Ours-Qwen3-4B} (\textit{right})}
        \label{tsne2D_model_y_L13}
    \end{subfigure}
    \vspace{0.01cm}

    \begin{subfigure}{\textwidth}
        \centering
        \includegraphics[width=0.8\textwidth]{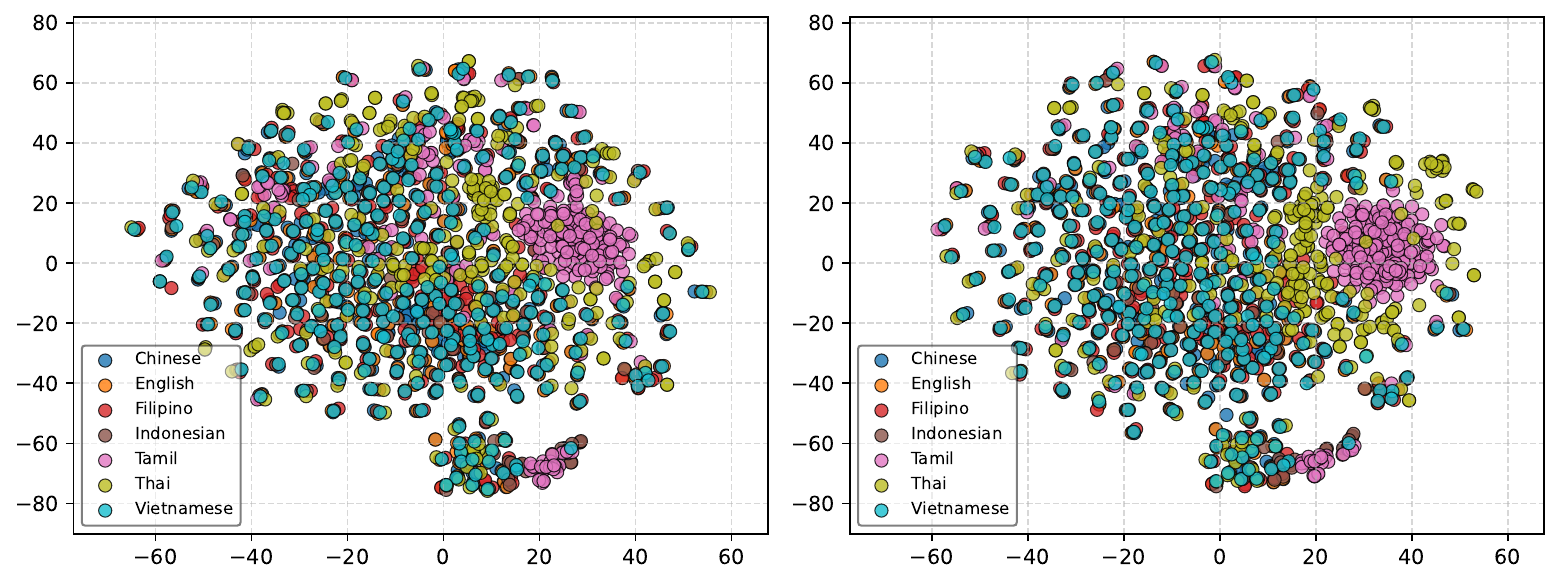}
        \caption{\texttt{Qwen3-VL-8B-Thinking} (\textit{left}) and \texttt{Ours-Qwen3-VL-8B} (\textit{right})}
        \label{tsne2D_model_z_L13}
    \end{subfigure}
    \vspace{0.01cm}

    \captionsetup{justification=centering}
    \caption{2D t-SNE visualizations of the middle layer (\textit{Layer 13}) multilingual representations for (a) \texttt{SmolLM3-3B}, (b) \texttt{Qwen3-4B-Thinking-2507}, and (c) \texttt{Qwen3-VL-8B-Thinking} before (\textit{left}) and after (\textit{right}) post-training with OSCD.}
    \label{tsne2D_L13}
\end{figure*}

\begin{figure*}[t] 
    \centering
    
    \begin{subfigure}{\textwidth}
        \centering
        \includegraphics[width=0.8\textwidth]{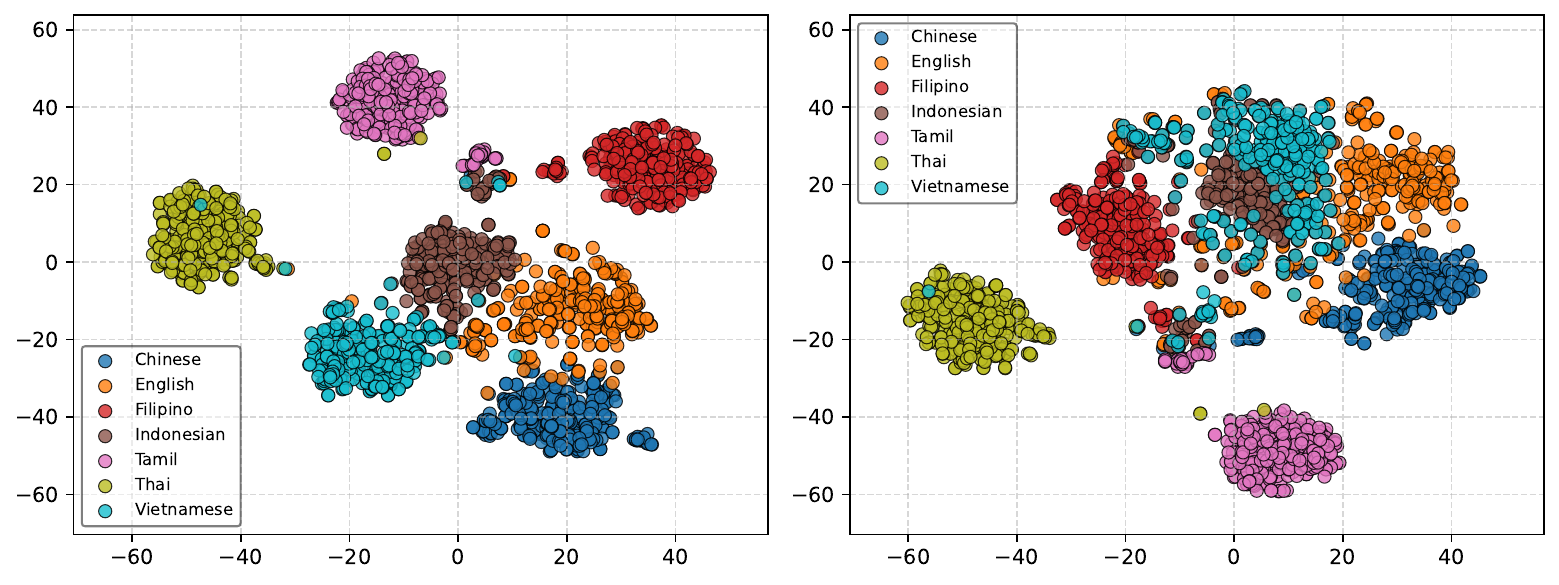}
        \caption{\texttt{SmolLM3-3B} (\textit{left}) and \texttt{Ours-SmolLM3-3B} (\textit{right})}
        \label{tsne2D_model_x_L25}
    \end{subfigure}
    \vspace{0.01cm} 

    \begin{subfigure}{\textwidth}
        \centering
        \includegraphics[width=0.8\textwidth]{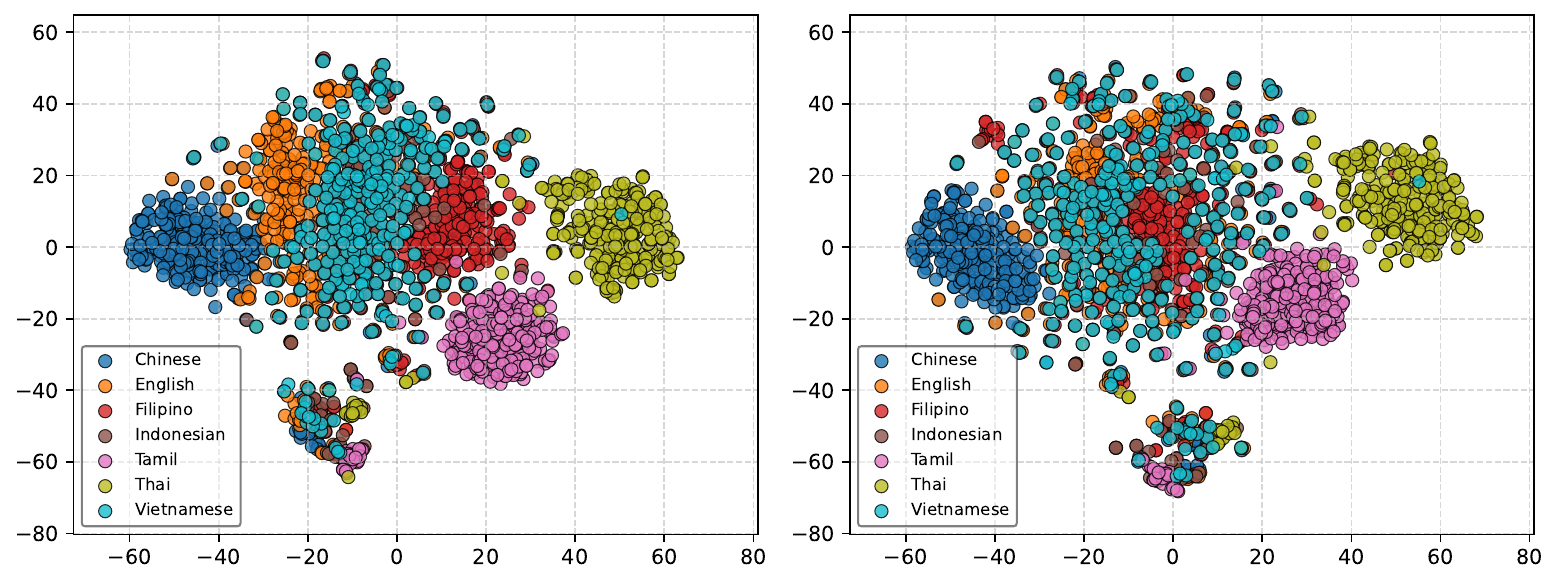}
        \caption{\texttt{Qwen3-4B-Thinking-2507} (\textit{left}) and \texttt{Ours-Qwen3-4B} (\textit{right})}
        \label{tsne2D_model_y_L25}
    \end{subfigure}
    \vspace{0.01cm}

    \begin{subfigure}{\textwidth}
        \centering
        \includegraphics[width=0.8\textwidth]{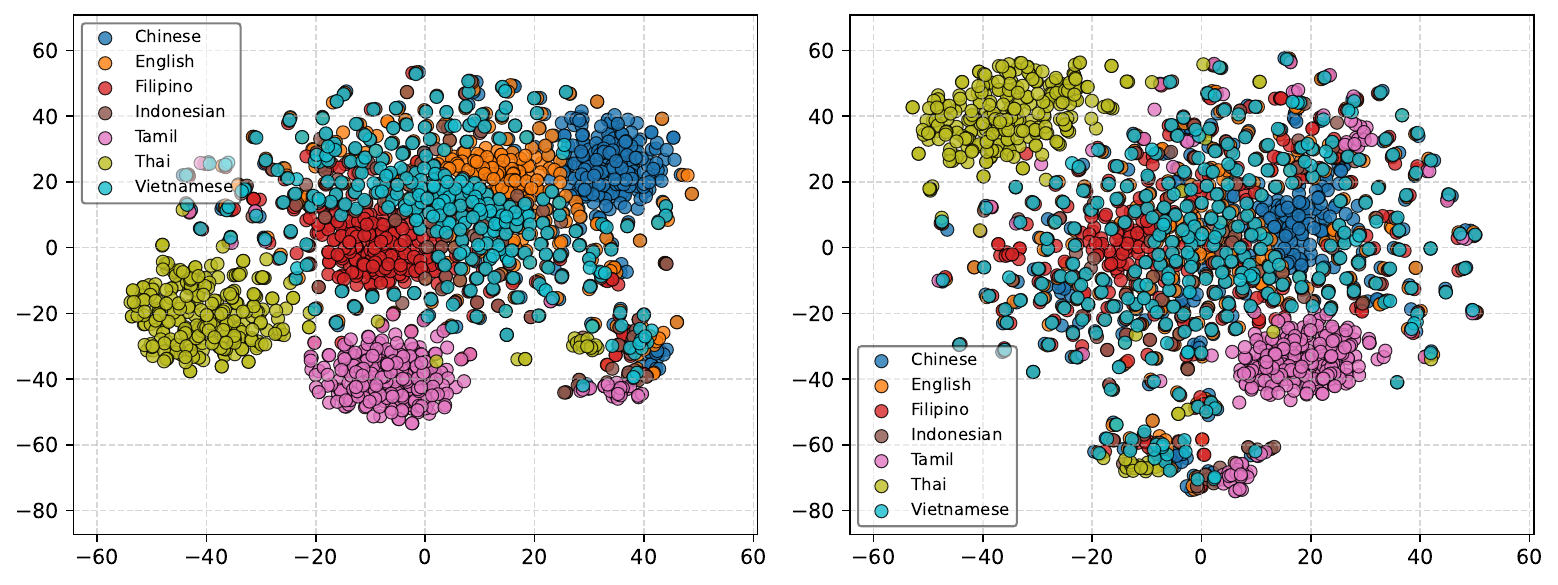}
        \caption{\texttt{Qwen3-VL-8B-Thinking} (\textit{left}) and \texttt{Ours-Qwen3-VL-8B} (\textit{right})}
        \label{tsne2D_model_z_L25}
    \end{subfigure}
    \vspace{0.01cm}

    \captionsetup{justification=centering}
    \caption{2D t-SNE visualizations of the middle layer (\textit{Layer 25}) multilingual representations for (a) \texttt{SmolLM3-3B}, (b) \texttt{Qwen3-4B-Thinking-2507}, and (c) \texttt{Qwen3-VL-8B-Thinking} before (\textit{left}) and after (\textit{right}) post-training with OSCD.}
    \label{tsne2D_L25}
\end{figure*}

\begin{figure*}[t]  
    \centering
    
    \begin{subfigure}{\textwidth}
        \centering
        \includegraphics[width=0.8\textwidth]{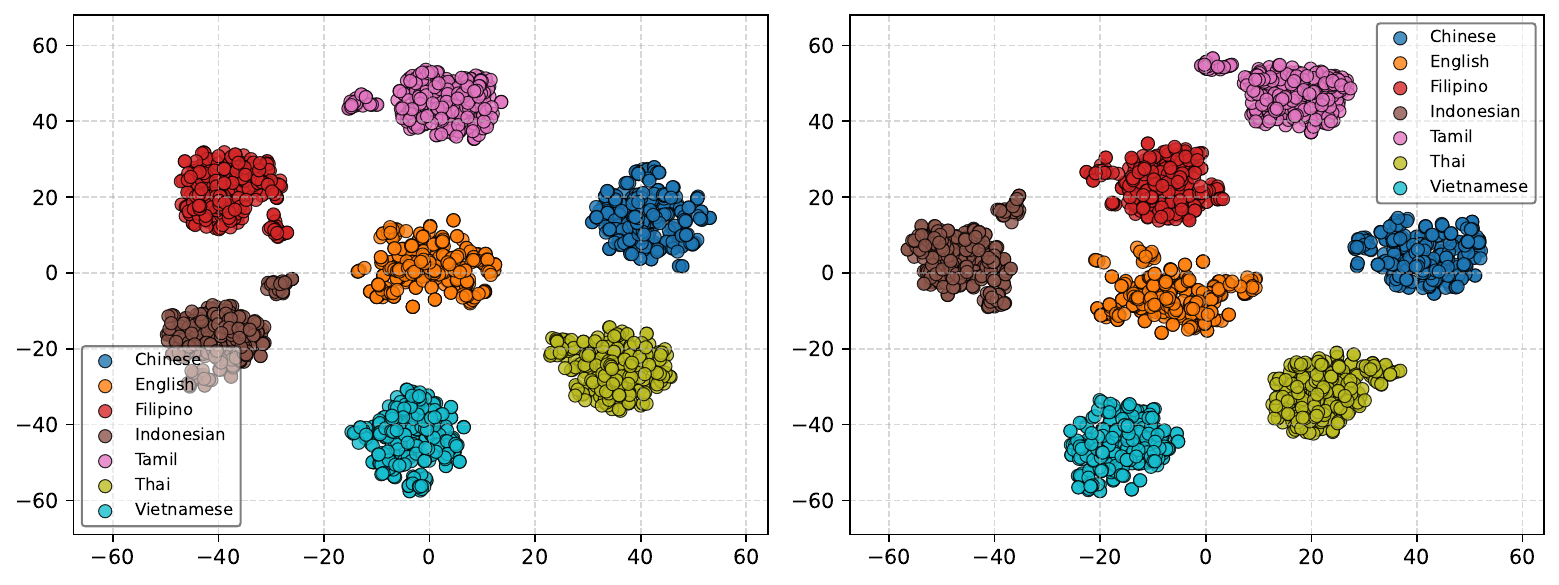}
        \caption{\texttt{SmolLM3-3B} (\textit{left}) and \texttt{Ours-SmolLM3-3B} (\textit{right})}
        \label{tsne2D_model_x_L36}
    \end{subfigure}
    \vspace{0.01cm} 

    \begin{subfigure}{\textwidth}
        \centering
        \includegraphics[width=0.8\textwidth]{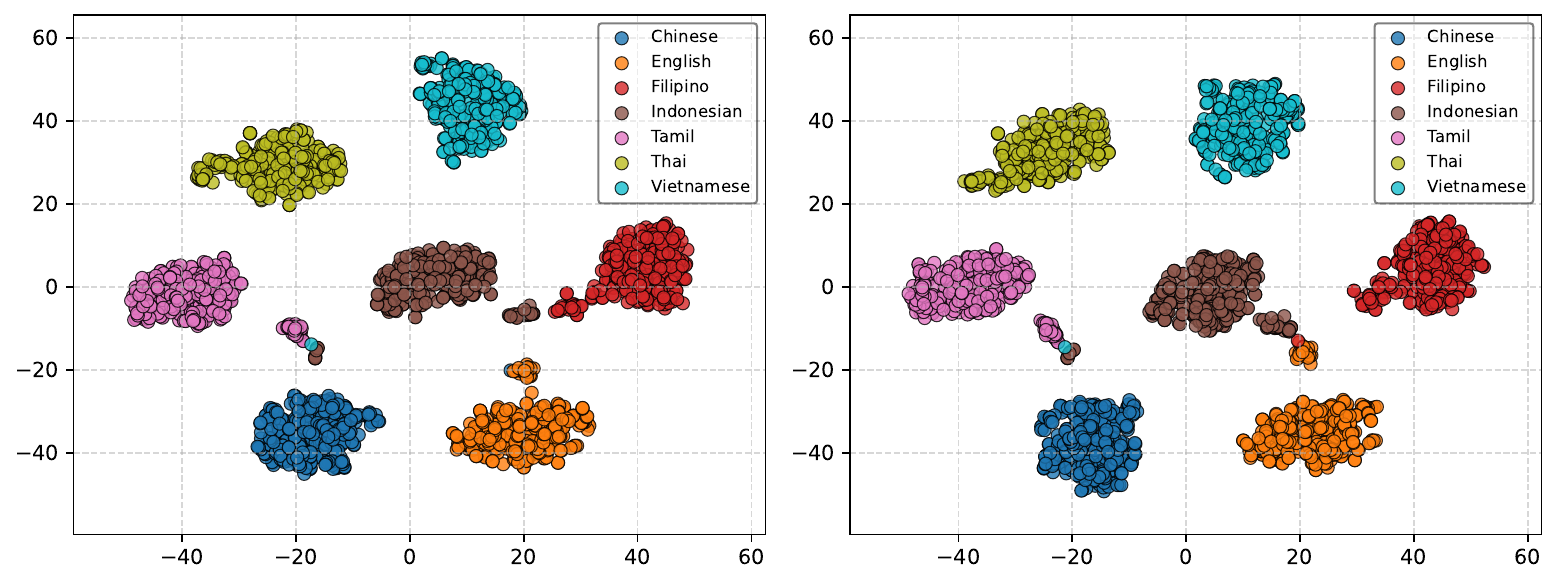}
        \caption{\texttt{Qwen3-4B-Thinking-2507} (\textit{left}) and \texttt{Ours-Qwen3-4B} (\textit{right})}
        \label{tsne2D_model_y_L36}
    \end{subfigure}
    \vspace{0.01cm}

    \begin{subfigure}{\textwidth}
        \centering
        \includegraphics[width=0.8\textwidth]{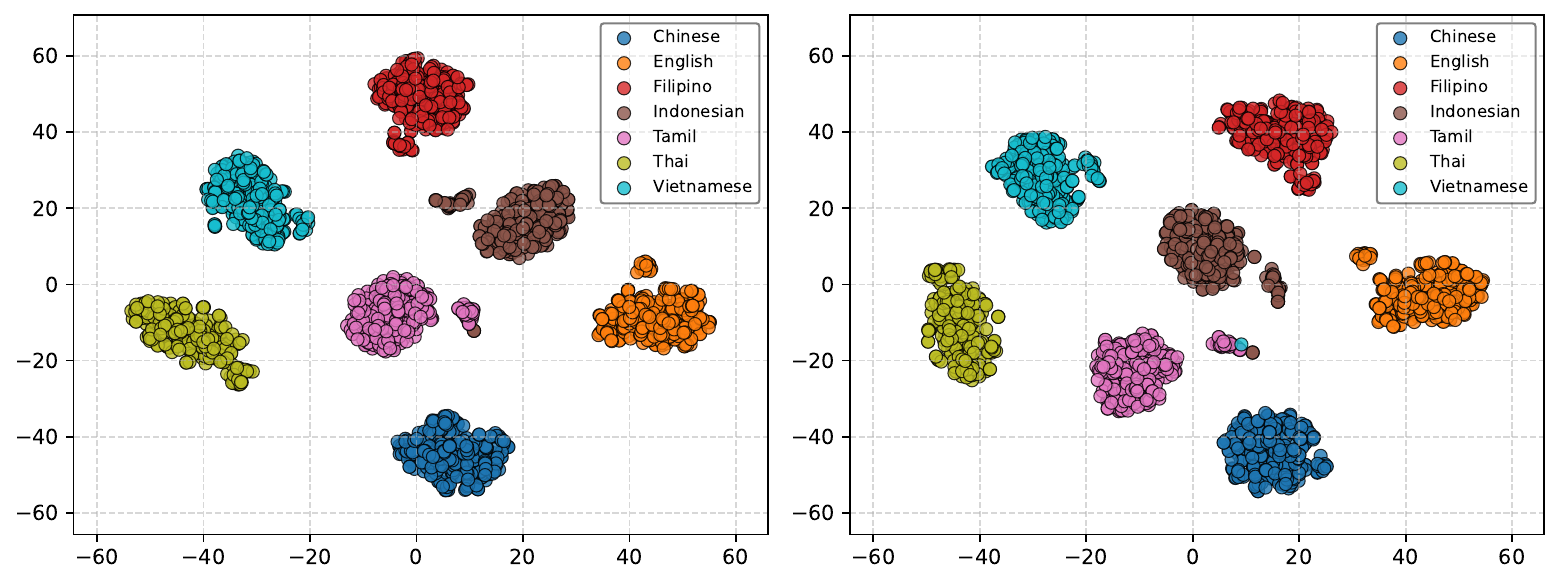}
        \caption{\texttt{Qwen3-VL-8B-Thinking} (\textit{left}) and \texttt{Ours-Qwen3-VL-8B} (\textit{right})}
        \label{tsne2D_model_z_L36}
    \end{subfigure}
    \vspace{0.01cm}

    \captionsetup{justification=centering}
    \caption{2D t-SNE visualizations of the last layer (\textit{Layer 36}) multilingual representations for (a) \texttt{SmolLM3-3B}, (b) \texttt{Qwen3-4B-Thinking-2507}, and (c) \texttt{Qwen3-VL-8B-Thinking} before (\textit{left}) and after (\textit{right}) post-training with OSCD.}
    \label{tsne2D_L36}
\end{figure*}

We provide the hidden-state representations of models before and after OSCD post-training using t-distributed Stochastic Neighbor Embedding (t-SNE) visualizations.
Figures~\ref{tsne2D_L1}--\ref{tsne2D_L36} illustrate the 2D t-SNE projections across sequential layers, 
using 300 parallel texts spanning the 7 target SEA languages, subsampled from the \texttt{FLORES-Plus} \textit{devtest} split.

\clearpage 
\begin{algorithm*}[t]  
\small
\caption{Onramp Sequence Cross-Distillation}
\label{cot_mutation_distill}
\begin{algorithmic}[1]
\REQUIRE Student Model $\pi_\theta$; Reference Model $\pi_\phi$; Translator System $\mathcal{T}$ 
\REQUIRE High-Resource Base Languages $\mathcal{M} = \{\text{EN}, \text{ZH}\}$; Escape Tokens $\Delta = \{t_{\texttt{<think>}}, t_{\texttt{</think>}}\}$ 
\REQUIRE Low-Resource Prompt-Language Pairs $\mathcal{P} = \{(p_i, l_i)\}_{i=1}^B$, where $\mathcal{L} = \{l_i\}_{i=1}^B$ and $\mathcal{L} \cap \mathcal{M} = \emptyset$
\REQUIRE Loss Weights: $\lambda_{\text{CE}}=1.0, \lambda_{\text{JEPA}}=1.0$

\FOR{each training step}

    \STATE \textbf{1. Rollout and Modify Reasoning}
    \STATE Sample reference completion IDs $c_i \sim \pi_\phi(\cdot \mid p_i)$ for $i \in [1, B]$  
    \FOR{$i = 1$ \textbf{to} $B$}
        \STATE \algorithmicif\ $l_i \in \mathcal{M}$ \algorithmicthen\ \textbf{continue}
        \STATE $ \text{str\_}r_i, \text{str\_}a_i\ \leftarrow \text{split}\big(\pi_\theta.\text{decode}(c_i), \text{delimiters}=\{\text{decode}(t) \mid t \in \Delta\}\big)$
        \STATE $\text{str\_}r'_i \leftarrow \mathcal{T}.\text{modify}(\text{str\_}r_i, \text{target\_lang}=l_i)$ 
        \hfill $\triangleright$ Translate reasoning trace
        \STATE $\mathbf{tok}_{r'_i}, \mathbf{tok}_{a_i} \leftarrow \pi_\theta.\text{encode}(\{ \text{str\_}r'_i, \text{str\_}a_i \}, \text{add\_special\_tokens=False})$ 
        \STATE $c'_i \leftarrow (t_{\texttt{<think>}} \text{ if } t_{\texttt{<think>}} \in c_i \text{ else } \emptyset) \oplus \mathbf{tok}_{r'_i} \oplus t_{\texttt{</think>}} \oplus \mathbf{tok}_{a_i}$ 
        \hfill $\triangleright$ Reconstruct completion IDs
    \ENDFOR 

    \STATE \textbf{2. Multi-Objective Loss Computation}
    \STATE Using TrainEngine, compute: 
    \STATE $\quad$ Logprobs $P^T$ and hidden states $H^T$ for $c_i$ given $\pi_\phi$, where $k = \text{pos}(t_{\texttt{</think>}} \in c_i)$  
    \STATE $\quad$ Logprobs $P^S$ and hidden states $H^S$ for $c'_i$ given $\pi_\theta$, where $k' = \text{pos}(t_{\texttt{</think>}} \in c'_i)$  

    \FOR{$i = 1$ \textbf{to} $B$}
        \STATE \textbf{Cross-Entropy}
        \STATE $\mathcal{L}_{\text{CE}}^{(i)} \leftarrow - \frac{1}{|c'_i|} \sum_{t=0}^{|c'_i|-1} \log P^S(c'_{i,t} \mid p_i, c'_{i,<t})$
        \hfill $\triangleright$ Completion w/ modified reasoning

        \STATE \textbf{Cosine Distance}
        \STATE $h^S_i, h^T_i \leftarrow H^S_{i=k'}, H^T_{i=k}$ 
        \STATE $\mathcal{L}_{\text{JEPA}}^{(i)} \leftarrow 1 - \text{cos\_sim}(h^S_i, h^T_i)$ 
        \hfill $\triangleright$ Let $\text{cos\_sim}(u,v) = \frac{u \cdot v}{\|u\|\|v\|}$
    
    \ENDFOR

    \STATE \textbf{3. Gradient Computation and Update}
    \STATE $\mathcal{L}_{\text{total}} \leftarrow \frac{1}{B} \sum_{i=1}^B \left( \lambda_{\text{CE}} \mathcal{L}_{\text{CE}}^{(i)} + \lambda_{\text{JEPA}} \mathcal{L}_{\text{JEPA}}^{(i)} \right)$
    \STATE Update parameters: $\theta \leftarrow \theta - \eta \nabla_\theta \mathcal{L}_{\text{total}}$

\ENDFOR
\end{algorithmic}
\end{algorithm*}

\begin{table*}[t]
\small
\centering
\captionsetup{justification=raggedright, singlelinecheck=false}
\begin{tabularx}{\textwidth}{X}
\toprule
\textbf{Prompt Template for Linguistic Classification} \\
\midrule
\textbf{[System]} \\
You are an impartial linguistic judge. Evaluate the provided text and determine the primary language utilized for its core communicative framework. \\
\\
\textbf{Evaluation Guidelines:} \\
\begin{enumerate}[leftmargin=*, noitemsep, topsep=2pt]
    \item \textbf{Structural Focus:} Focus exclusively on syntax, grammatical connectors, and structural framework. Ignore target object bias such as the specific data being analyzed, translated, or discussed.
    \item \textbf{Candidate Labels:} Classify the language strictly into one of the following permissible categories: \{\texttt{English}, \texttt{Chinese}, \texttt{Filipino}, \texttt{Indonesian}, \texttt{Tamil}, \texttt{Thai}, \texttt{Vietnamese}\}.
    \item \textbf{Cross-Tier Resolution:} If language-switching or multi-lingual mixing occurs between Tier 1 (English, Chinese) and Tier 2 (other languages), instantly disqualify Tier 2 and default to the Tier 1 language.
    \item \textbf{Intra-Tier Resolution:} If both Tier 1 languages (English and Chinese) co-occur anywhere within the text, default strictly to English as the final structural fallback category.
\end{enumerate} \\\\
Begin the evaluation with a structured linguistic breakdown separating framework grammar from referenced data. If mixing occurred, explicitly justify the tier-resolution application. \\
\\
Conclude the response \textbf{STRICTLY} in this format: ``Result: \textbackslash boxed\{Language\}''. \\
\\
\textbf{[Input Text]} \\
\texttt{\{\{\texttt{REASONING\_TRACE}\}\}} \\\\
\textbf{[Evaluation]} \\
\bottomrule
\end{tabularx}
\caption{Prompt template for LLM-as-a-judge classification of \texttt{REASONING\_TRACE} language.}
\label{classification_prompt}
\end{table*}

\begin{table*}[t]
\small
\centering
\captionsetup{justification=raggedright, singlelinecheck=false}
\begin{tabularx}{\textwidth}{X}
\toprule
\textbf{Prompt Template for Zero-Shot Translation} \\
\midrule
\textbf{[System]} \\
You are a professional translator. Your primary job is to translate the provided text from its original source language into the designated target language: \texttt{\{\{TARGET\_LANG\}\}}. \\
\\
\textbf{Translation Guidelines:} \\
\begin{enumerate}[leftmargin=*, noitemsep, topsep=2pt]
    \item \textbf{Accuracy \& Tone:} Preserve the exact meaning, context, and original tone (formal, casual, technical, etc.). Maintain all semantic nuances without omission.
    \item \textbf{Localization:} Adapt cultural references, idioms, numerical formats, dates, and currencies to native target language conventions seamlessly.
    \item \textbf{Fluency \& Formatting:} Ensure a natural, coherent flow. Maintain original formatting (bold, italics) and preserve proper nouns unless standard translations exist.
    \item \textbf{Incomplete Input:} If the text is truncated or ends mid-sentence, use the preceding context to logically and structurally complete the translation.
\end{enumerate} \\\\
Return \textbf{ONLY} the finished translation. \textbf{STRICTLY DO NOT} include any meta-comments, placeholders, introductory text, or explanations. \\
\\
\textbf{[Original Text]} \\
\texttt{\{\{SOURCE\_TEXT\}\}} \\\\
\textbf{[Translated Text]} \\
\bottomrule
\end{tabularx}
\caption{Prompt template for zero-shot translation of \texttt{SOURCE\_TEXT} from source language to \texttt{TARGET\_LANG}.}
\label{translation_prompt}
\end{table*}

\begin{table*}[t]
\small
\centering
\captionsetup{justification=raggedright, singlelinecheck=false}
\begin{tabularx}{\textwidth}{X}
\toprule
\textbf{Prompt Template for Parallel Text Evaluation} \\
\midrule
\textbf{[System]} \\
You are an impartial translation judge. Your primary job is to evaluate the quality of the provided translation based on accuracy, fluency, and adherence to structural constraints. \\
\\
\textbf{Evaluation Guidelines:} \\
\begin{enumerate}[leftmargin=*, noitemsep, topsep=2pt]
    \item \textbf{Scoring Metric:} Rate the translation on an integer scale from 1 to 10 using a standard rubric. Assess semantic accuracy, nuance preservation, and intent alignment.
    \item \textbf{Language Homogeneity:} Ensure the translation remains completely homogeneous. Technical jargon or specific domain terminology may remain untranslated only if core logical connectors maintain the target framework.
    \item \textbf{Leakage Penalty:} If the translation contains language leakage (e.g., untranslated prose, mixed-language sentences) violating homogeneity, automatically award a strict final score of 1.
    \item \textbf{Format Compliance:} Verify that the output strictly adheres to the requested markdown conventions, structural alignment, and tag parameters without introducing peripheral text.
\end{enumerate} \\\\
Begin the evaluation with a concise explanation detailing the specific strengths and weaknesses observed based on the rubric criteria. \\
\\
Conclude the response \textbf{STRICTLY} in this format: ``Result: \textbackslash boxed\{Score\}''. \\
\\
\textbf{[Original Text]} \\
\texttt{\{\{\texttt{ORIGINAL\_REASONING\_TRACE}\}\}} \\\\
\textbf{[Translated Text]} \\
\texttt{\{\{\texttt{TRANSLATED\_REASONING\_TRACE}\}\}} \\\\
\textbf{[Evaluation]} \\
\bottomrule
\end{tabularx}
\caption{Prompt template for LLM-as-a-judge quality evaluation of parallel text \texttt{REASONING\_TRACEs}.}
\label{evaluation_prompt}
\end{table*}

\clearpage 

\begin{table*}[t]
\centering
\small
\setlength{\tabcolsep}{0pt} 
\begin{tabular*}{\textwidth}{l @{\extracolsep{\fill}} ccc >{\columncolor[gray]{0.85}}c ccc >{\columncolor[gray]{0.85}}c }
\hline
& \multicolumn{4}{c}{\textbf{Any-CoT}} & \multicolumn{4}{c}{\textbf{Target-CoT}} \\
\cline{2-5} \cline{6-9}
\textbf{Model} & \textbf{Chinese} & \textbf{English} & \textbf{Indonesian} & \textbf{Overall} & \textbf{Chinese} & \textbf{English} & \textbf{Indonesian} & \textbf{Overall} \\
\hline
\textit{Pass@$5$ (\%) $\uparrow$} \\
Qwen-SEA-LION-v4-8B-VL & 53.3 & 63.3 & 50.0 & 55.6 {\scalebox{0.8}{$\pm$6.9}} & 36.7 & 63.3 & 43.3 & 47.8 {\scalebox{0.8}{$\pm$13.9}} \\
Qwen3-VL-8B-Thinking & 73.3 & \textbf{\color{red}86.7} & \textbf{\color{red}86.7} & \textbf{\color{red}82.2 {\scalebox{0.8}{$\pm$7.7}}} & 63.3 & \textbf{\color{red}86.7} & 0.0 & 50.0 {\scalebox{0.8}{$\pm$44.8}} \\
\textbf{Ours-Qwen3-8B-$\mathcal{L}_{\text{CE}}$} & 46.7 & 73.3 & 50.0 & 56.7 {\scalebox{0.8}{$\pm$14.5}} & 23.3 & 73.3 & 16.7 & 37.8 {\scalebox{0.8}{$\pm$31.0}} \\
\textbf{Ours-Qwen3-8B-$\mathcal{L}_{\text{CE+JEPA}}$} & 43.3 & 70.0 & 63.3 & 58.9 {\scalebox{0.8}{$\pm$13.9}} & 26.7 & 70.0 & 23.3 & 40.0 {\scalebox{0.8}{$\pm$26.0}} \\
\textbf{Ours-Qwen3-8B-$\mathcal{L}_{\text{CE (Agentic)}}$} & \textbf{\color{red}76.7} & 83.3 & 80.0 & 80.0 {\scalebox{0.8}{$\pm$3.3}} & 66.7 & 80.0 & \textbf{\color{red}76.7} & 74.4 {\scalebox{0.8}{$\pm$6.9}} \\
\textbf{Ours-Qwen3-8B-$\mathcal{L}_{\text{CE+JEPA (Agentic)}}$} & \textbf{\color{red}76.7} & 83.3 & 80.0 & 80.0 {\scalebox{0.8}{$\pm$3.3}} & \textbf{\color{red}76.7} & 83.3 & \textbf{\color{red}76.7} & \textbf{\color{red}78.9 {\scalebox{0.8}{$\pm$3.8}}} \\ \\\\[0.5ex]
\textit{Mean@$5$ (\%) $\pm$ std $\uparrow$} \\
Qwen-SEA-LION-v4-8B-VL & 36.7 {\scalebox{0.8}{$\pm$4.2}} & 40.7 {\scalebox{0.8}{$\pm$3.9}} & 34.7 {\scalebox{0.8}{$\pm$3.4}} & 37.3 {\scalebox{0.8}{$\pm$3.1}} & 19.3 {\scalebox{0.8}{$\pm$6.5}} & 38.7 {\scalebox{0.8}{$\pm$5.4}} & 27.3 {\scalebox{0.8}{$\pm$2.5}} & 28.4 {\scalebox{0.8}{$\pm$9.7}} \\
Qwen3-VL-8B-Thinking & \textbf{\color{red}61.3 {\scalebox{0.8}{$\pm$4.5}}} & \textbf{\color{red}78.7 {\scalebox{0.8}{$\pm$3.4}}} & \textbf{\color{red}68.7 {\scalebox{0.8}{$\pm$5.4}}} & \textbf{\color{red}69.6 {\scalebox{0.8}{$\pm$8.7}}} & \textbf{\color{red}50.0 {\scalebox{0.8}{$\pm$2.1}}} & \textbf{\color{red}74.7 {\scalebox{0.8}{$\pm$3.4}}} & 0.0 {\scalebox{0.8}{$\pm$0.0}} & 41.6 {\scalebox{0.8}{$\pm$38.0}} \\
\textbf{Ours-Qwen3-8B-$\mathcal{L}_{\text{CE}}$} & 26.7 {\scalebox{0.8}{$\pm$8.2}} & 55.3 {\scalebox{0.8}{$\pm$4.5}} & 28.7 {\scalebox{0.8}{$\pm$5.8}} & 36.9 {\scalebox{0.8}{$\pm$16.0}} & 13.3 {\scalebox{0.8}{$\pm$4.7}} & 51.7 {\scalebox{0.8}{$\pm$3.7}} & 5.0 {\scalebox{0.8}{$\pm$3.7}} & 23.3 {\scalebox{0.8}{$\pm$24.9}} \\
\textbf{Ours-Qwen3-8B-$\mathcal{L}_{\text{CE+JEPA}}$} & 25.3 {\scalebox{0.8}{$\pm$5.0}} & 54.7 {\scalebox{0.8}{$\pm$6.2}} & 28.0 {\scalebox{0.8}{$\pm$7.2}} & 36.0 {\scalebox{0.8}{$\pm$16.2}} & 9.2 {\scalebox{0.8}{$\pm$2.8}} & 55.0 {\scalebox{0.8}{$\pm$3.7}} & 8.3 {\scalebox{0.8}{$\pm$3.7}} & 24.2 {\scalebox{0.8}{$\pm$26.7}} \\
\textbf{Ours-Qwen3-8B-$\mathcal{L}_{\text{CE (Agentic)}}$} & 58.0 {\scalebox{0.8}{$\pm$6.2}} & 72.7 {\scalebox{0.8}{$\pm$4.4}} & 57.3 {\scalebox{0.8}{$\pm$10.2}} & 62.7 {\scalebox{0.8}{$\pm$8.7}} & 42.5 {\scalebox{0.8}{$\pm$6.4}} & 68.3 {\scalebox{0.8}{$\pm$5.5}} & \textbf{\color{red}47.5 {\scalebox{0.8}{$\pm$10.1}}} & 52.8 {\scalebox{0.8}{$\pm$13.7}} \\
\textbf{Ours-Qwen3-8B-$\mathcal{L}_{\text{CE+JEPA (Agentic)}}$} & 59.3 {\scalebox{0.8}{$\pm$6.5}} & 70.7 {\scalebox{0.8}{$\pm$2.5}} & 63.3 {\scalebox{0.8}{$\pm$6.0}} & 64.4 {\scalebox{0.8}{$\pm$5.7}} & 48.3 {\scalebox{0.8}{$\pm$5.0}} & 67.5 {\scalebox{0.8}{$\pm$4.3}} & 46.7 {\scalebox{0.8}{$\pm$2.4}} & \textbf{\color{red}54.2 {\scalebox{0.8}{$\pm$11.6}}} \\
\hline
\end{tabular*}
\caption{\label{aime25-ablation-micro}
Ablation performance on the \texttt{AIME25} benchmark comparing different training loss configurations,
evaluating Any-CoT vs Target-CoT setups across 3 target languages (\texttt{ZH,EN,IN}). 
}
\end{table*}

\begin{table*}[t]
\centering
\small
\setlength{\tabcolsep}{0pt} 
\begin{tabular*}{\textwidth}{l @{\extracolsep{\fill}} ccc >{\columncolor[gray]{0.85}}c ccc >{\columncolor[gray]{0.85}}c }
\hline
& \multicolumn{4}{c}{\textbf{Any-CoT}} & \multicolumn{4}{c}{\textbf{Target-CoT}} \\
\cline{2-5} \cline{6-9}
\textbf{Model} & \textbf{Chinese} & \textbf{English} & \textbf{Indonesian} & \textbf{Overall} & \textbf{Chinese} & \textbf{English} & \textbf{Indonesian} & \textbf{Overall} \\
\hline
\textit{Pass@$5$ (\%) $\uparrow$} \\
Qwen-SEA-LION-v4-8B-VL & 43.3 & 40.0 & 30.0 & 37.8 {\scalebox{0.8}{$\pm$6.9}} & 33.3 & 40.0 & 23.3 & 32.2 {\scalebox{0.8}{$\pm$8.4}} \\
Qwen3-VL-8B-Thinking & 43.3 & 56.7 & \textbf{\color{red}56.7} & \textbf{\color{red}52.2 {\scalebox{0.8}{$\pm$7.7}}} & 40.0 & \textbf{\color{red}56.7} & 0.0 & 32.2 {\scalebox{0.8}{$\pm$29.1}} \\
\textbf{Ours-Qwen3-8B-$\mathcal{L}_{\text{CE}}$} & 33.3 & \textbf{\color{red}60.0} & 33.3 & 42.2 {\scalebox{0.8}{$\pm$15.4}} & 23.3 & \textbf{\color{red}56.7} & 13.3 & 31.1 {\scalebox{0.8}{$\pm$22.7}} \\
\textbf{Ours-Qwen3-8B-$\mathcal{L}_{\text{CE+JEPA}}$} & 26.7 & 46.7 & 33.3 & 35.6 {\scalebox{0.8}{$\pm$10.2}} & 23.3 & 46.7 & 13.3 & 27.8 {\scalebox{0.8}{$\pm$17.1}} \\
\textbf{Ours-Qwen3-8B-$\mathcal{L}_{\text{CE (Agentic)}}$} & \textbf{\color{red}50.0} & 56.7 & 43.3 & 50.0 {\scalebox{0.8}{$\pm$6.7}} & 46.7 & \textbf{\color{red}56.7} & 40.0 & 47.8 {\scalebox{0.8}{$\pm$8.4}} \\
\textbf{Ours-Qwen3-8B-$\mathcal{L}_{\text{CE+JEPA (Agentic)}}$} & \textbf{\color{red}50.0} & 56.7 & 46.7 & 51.1 {\scalebox{0.8}{$\pm$5.1}} & \textbf{\color{red}50.0} & 53.3 & \textbf{\color{red}43.3} & \textbf{\color{red}48.9 {\scalebox{0.8}{$\pm$5.1}}} \\ \\\\[0.5ex]
\textit{Mean@$5$ (\%) $\pm$ std $\uparrow$} \\
Qwen-SEA-LION-v4-8B-VL & 26.0 {\scalebox{0.8}{$\pm$4.9}} & 26.7 {\scalebox{0.8}{$\pm$4.2}} & 20.0 {\scalebox{0.8}{$\pm$2.1}} & 24.2 {\scalebox{0.8}{$\pm$3.7}} & 14.7 {\scalebox{0.8}{$\pm$3.4}} & 23.3 {\scalebox{0.8}{$\pm$5.6}} & 15.3 {\scalebox{0.8}{$\pm$1.6}} & 17.8 {\scalebox{0.8}{$\pm$4.8}} \\
Qwen3-VL-8B-Thinking & 38.0 {\scalebox{0.8}{$\pm$4.0}} & \textbf{\color{red}50.0 {\scalebox{0.8}{$\pm$4.7}}} & \textbf{\color{red}45.3 {\scalebox{0.8}{$\pm$3.4}}} & \textbf{\color{red}44.4 {\scalebox{0.8}{$\pm$6.0}}} & \textbf{\color{red}28.0 {\scalebox{0.8}{$\pm$4.5}}} & \textbf{\color{red}50.0 {\scalebox{0.8}{$\pm$4.7}}} & 0.0 {\scalebox{0.8}{$\pm$0.0}} & 26.0 {\scalebox{0.8}{$\pm$25.1}} \\
\textbf{Ours-Qwen3-8B-$\mathcal{L}_{\text{CE}}$} & 20.0 {\scalebox{0.8}{$\pm$6.3}} & 39.3 {\scalebox{0.8}{$\pm$2.5}} & 22.7 {\scalebox{0.8}{$\pm$2.5}} & 27.3 {\scalebox{0.8}{$\pm$10.5}} & 11.7 {\scalebox{0.8}{$\pm$5.0}} & 38.3 {\scalebox{0.8}{$\pm$1.7}} & 7.5 {\scalebox{0.8}{$\pm$1.4}} & 19.2 {\scalebox{0.8}{$\pm$16.7}} \\
\textbf{Ours-Qwen3-8B-$\mathcal{L}_{\text{CE+JEPA}}$} & 15.3 {\scalebox{0.8}{$\pm$4.0}} & 32.0 {\scalebox{0.8}{$\pm$2.7}} & 16.0 {\scalebox{0.8}{$\pm$4.9}} & 21.1 {\scalebox{0.8}{$\pm$9.4}} & 9.2 {\scalebox{0.8}{$\pm$4.9}} & 32.5 {\scalebox{0.8}{$\pm$2.8}} & 3.3 {\scalebox{0.8}{$\pm$2.4}} & 15.0 {\scalebox{0.8}{$\pm$15.4}} \\
\textbf{Ours-Qwen3-8B-$\mathcal{L}_{\text{CE (Agentic)}}$} & \textbf{\color{red}40.7 {\scalebox{0.8}{$\pm$3.3}}} & 40.7 {\scalebox{0.8}{$\pm$1.3}} & 31.3 {\scalebox{0.8}{$\pm$1.6}} & 37.6 {\scalebox{0.8}{$\pm$5.4}} & 23.3 {\scalebox{0.8}{$\pm$8.5}} & 40.8 {\scalebox{0.8}{$\pm$1.4}} & 20.8 {\scalebox{0.8}{$\pm$6.0}} & 28.3 {\scalebox{0.8}{$\pm$10.9}} \\
\textbf{Ours-Qwen3-8B-$\mathcal{L}_{\text{CE+JEPA (Agentic)}}$} & 37.3 {\scalebox{0.8}{$\pm$3.3}} & 45.3 {\scalebox{0.8}{$\pm$3.4}} & 35.3 {\scalebox{0.8}{$\pm$4.0}} & 39.3 {\scalebox{0.8}{$\pm$5.3}} & 26.7 {\scalebox{0.8}{$\pm$7.8}} & 42.5 {\scalebox{0.8}{$\pm$2.8}} & \textbf{\color{red}26.7 {\scalebox{0.8}{$\pm$2.4}}} & \textbf{\color{red}31.9 {\scalebox{0.8}{$\pm$9.1}}} \\
\hline
\end{tabular*}
\caption{\label{hmmt25-ablation-micro}
Ablation performance on the \texttt{HMMT25} benchmark comparing different training loss configurations,
evaluating Any-CoT vs Target-CoT setups across 3 target languages (\texttt{ZH,EN,IN}). 
}
\end{table*}

\clearpage 

\begin{table*}[t]
\centering
\small
\setlength{\tabcolsep}{0pt} 
\begin{tabular*}{\textwidth}{l @{\extracolsep{\fill}} ccccccc >{\columncolor[gray]{0.85}}c }
\hline
& \multicolumn{8}{c}{\textbf{Any-CoT}} \\
\cline{2-9}
\textbf{Model} & \textbf{Chinese} & \textbf{English} & \textbf{Filipino} & \textbf{Indonesian} & \textbf{Tamil} & \textbf{Thai} & \textbf{Vietnamese} & \textbf{Overall} \\
\hline
\textit{Pass@$5$ (\%) $\uparrow$} \\
SmolLM3-3B & 30.0 & 60.0 & 43.3 & 46.7 & 13.3 & 46.7 & 53.3 & 41.9 {\scalebox{0.8}{$\pm$15.6}} \\
\textbf{Ours-SmolLM3-3B} & 40.0 & 53.3 & 33.3 & 36.7 & 10.0 & 16.7 & 30.0 & 31.4 {\scalebox{0.8}{$\pm$14.5}} \\
Qwen3-4B-Thinking-2507 & 80.0 & \textbf{\color{red}86.7} & \textbf{\color{red}83.3} & \textbf{\color{red}86.7} & \textbf{\color{red}83.3} & 83.3 & 83.3 & 83.8 {\scalebox{0.8}{$\pm$2.3}} \\
\textbf{Ours-Qwen3-4B} & \textbf{\color{red}86.7} & \textbf{\color{red}86.7} & 80.0 & \textbf{\color{red}86.7} & \textbf{\color{red}83.3} & \textbf{\color{red}86.7} & \textbf{\color{red}86.7} & \textbf{\color{red}85.2 {\scalebox{0.8}{$\pm$2.6}}} \\
Qwen-SEA-LION-v4-8B-VL & 53.3 & 63.3 & 50.0 & 50.0 & 30.0 & 46.7 & 53.3 & 49.5 {\scalebox{0.8}{$\pm$10.1}} \\
Qwen3-VL-8B-Thinking & 73.3 & \textbf{\color{red}86.7} & \textbf{\color{red}83.3} & \textbf{\color{red}86.7} & \textbf{\color{red}83.3} & 83.3 & \textbf{\color{red}86.7} & 83.3 {\scalebox{0.8}{$\pm$4.7}} \\
\textbf{Ours-Qwen3-VL-8B} & 80.0 & 83.3 & 76.7 & 76.7 & 50.0 & 73.3 & 76.7 & 73.8 {\scalebox{0.8}{$\pm$11.0}} \\ \\\\[0.5ex]
\textit{Mean@$5$ (\%) $\pm$ std $\uparrow$} \\
SmolLM3-3B & 14.7 {\scalebox{0.8}{$\pm$2.7}} & 38.0 {\scalebox{0.8}{$\pm$2.7}} & 27.3 {\scalebox{0.8}{$\pm$5.7}} & 26.7 {\scalebox{0.8}{$\pm$4.2}} & 5.3 {\scalebox{0.8}{$\pm$1.6}} & 26.0 {\scalebox{0.8}{$\pm$6.8}} & 24.7 {\scalebox{0.8}{$\pm$5.4}} & 23.2 {\scalebox{0.8}{$\pm$10.4}} \\
\textbf{Ours-SmolLM3-3B} & 23.3 {\scalebox{0.8}{$\pm$3.7}} & 32.0 {\scalebox{0.8}{$\pm$6.9}} & 20.7 {\scalebox{0.8}{$\pm$3.3}} & 19.3 {\scalebox{0.8}{$\pm$2.5}} & 5.3 {\scalebox{0.8}{$\pm$2.7}} & 12.0 {\scalebox{0.8}{$\pm$1.6}} & 17.3 {\scalebox{0.8}{$\pm$2.5}} & 18.6 {\scalebox{0.8}{$\pm$8.4}} \\
Qwen3-4B-Thinking-2507 & 60.0 {\scalebox{0.8}{$\pm$3.0}} & \textbf{\color{red}80.7 {\scalebox{0.8}{$\pm$2.5}}} & 69.3 {\scalebox{0.8}{$\pm$4.4}} & \textbf{\color{red}73.3 {\scalebox{0.8}{$\pm$3.0}}} & 57.3 {\scalebox{0.8}{$\pm$5.3}} & 71.3 {\scalebox{0.8}{$\pm$5.4}} & 70.7 {\scalebox{0.8}{$\pm$2.5}} & 69.0 {\scalebox{0.8}{$\pm$8.0}} \\
\textbf{Ours-Qwen3-4B} & \textbf{\color{red}73.3 {\scalebox{0.8}{$\pm$4.7}}} & 72.7 {\scalebox{0.8}{$\pm$3.3}} & \textbf{\color{red}72.0 {\scalebox{0.8}{$\pm$4.0}}} & 66.0 {\scalebox{0.8}{$\pm$4.4}} & 56.0 {\scalebox{0.8}{$\pm$3.3}} & \textbf{\color{red}72.0 {\scalebox{0.8}{$\pm$1.6}}} & \textbf{\color{red}71.3 {\scalebox{0.8}{$\pm$4.5}}} & \textbf{\color{red}69.0 {\scalebox{0.8}{$\pm$6.2}}} \\
Qwen-SEA-LION-v4-8B-VL & 36.7 {\scalebox{0.8}{$\pm$4.2}} & 40.7 {\scalebox{0.8}{$\pm$3.9}} & 34.0 {\scalebox{0.8}{$\pm$5.7}} & 34.7 {\scalebox{0.8}{$\pm$3.4}} & 15.3 {\scalebox{0.8}{$\pm$3.4}} & 30.0 {\scalebox{0.8}{$\pm$5.6}} & 34.0 {\scalebox{0.8}{$\pm$4.9}} & 32.2 {\scalebox{0.8}{$\pm$8.1}} \\
Qwen3-VL-8B-Thinking & 61.3 {\scalebox{0.8}{$\pm$4.5}} & 78.7 {\scalebox{0.8}{$\pm$3.4}} & 71.3 {\scalebox{0.8}{$\pm$2.7}} & 68.7 {\scalebox{0.8}{$\pm$5.4}} & \textbf{\color{red}63.3 {\scalebox{0.8}{$\pm$7.0}}} & 66.0 {\scalebox{0.8}{$\pm$6.5}} & 70.7 {\scalebox{0.8}{$\pm$9.8}} & 68.6 {\scalebox{0.8}{$\pm$5.8}} \\
\textbf{Ours-Qwen3-VL-8B} & 64.7 {\scalebox{0.8}{$\pm$3.4}} & 74.7 {\scalebox{0.8}{$\pm$4.0}} & 64.0 {\scalebox{0.8}{$\pm$5.7}} & 66.7 {\scalebox{0.8}{$\pm$6.0}} & 32.0 {\scalebox{0.8}{$\pm$5.0}} & 58.7 {\scalebox{0.8}{$\pm$2.7}} & 65.3 {\scalebox{0.8}{$\pm$2.7}} & 60.9 {\scalebox{0.8}{$\pm$13.6}} \\
\hline
\end{tabular*}
\caption{\label{aime25-main-any}
Main performance on the \texttt{AIME25} benchmark evaluating Any-CoT,
comparing base reasoning models and OSCD post-trained models across 7 target languages (\texttt{ZH,EN,Fi,IN,TA,TH,VI}). 
}
\end{table*}

\begin{table*}[t]
\centering
\small
\setlength{\tabcolsep}{0pt} 
\begin{tabular*}{\textwidth}{l @{\extracolsep{\fill}} ccccccc >{\columncolor[gray]{0.85}}c }
\hline
& \multicolumn{8}{c}{\textbf{Any-CoT}} \\
\cline{2-9}
\textbf{Model} & \textbf{Chinese} & \textbf{English} & \textbf{Filipino} & \textbf{Indonesian} & \textbf{Tamil} & \textbf{Thai} & \textbf{Vietnamese} & \textbf{Overall} \\
\hline
\textit{Pass@$5$ (\%) $\uparrow$} \\
SmolLM3-3B & 16.7 & 33.3 & 23.3 & 26.7 & 10.0 & 26.7 & 23.3 & 22.9 {\scalebox{0.8}{$\pm$7.6}} \\
\textbf{Ours-SmolLM3-3B} & 16.7 & 30.0 & 13.3 & 13.3 & 0.0 & 6.7 & 13.3 & 13.3 {\scalebox{0.8}{$\pm$9.2}} \\
Qwen3-4B-Thinking-2507 & 50.0 & 56.7 & 50.0 & \textbf{\color{red}56.7} & 43.3 & \textbf{\color{red}56.7} & \textbf{\color{red}56.7} & 52.9 {\scalebox{0.8}{$\pm$5.2}} \\
\textbf{Ours-Qwen3-4B} & \textbf{\color{red}56.7} & 56.7 & \textbf{\color{red}56.7} & 53.3 & \textbf{\color{red}53.3} & 50.0 & \textbf{\color{red}56.7} & \textbf{\color{red}54.8 {\scalebox{0.8}{$\pm$2.6}}} \\
Qwen-SEA-LION-v4-8B-VL & 43.3 & 40.0 & 30.0 & 30.0 & 10.0 & 30.0 & 33.3 & 31.0 {\scalebox{0.8}{$\pm$10.7}} \\
Qwen3-VL-8B-Thinking & 43.3 & 56.7 & 53.3 & \textbf{\color{red}56.7} & 46.7 & 53.3 & \textbf{\color{red}56.7} & 52.4 {\scalebox{0.8}{$\pm$5.3}} \\
\textbf{Ours-Qwen3-VL-8B} & 43.3 & \textbf{\color{red}60.0} & 46.7 & 46.7 & 20.0 & 40.0 & 50.0 & 43.8 {\scalebox{0.8}{$\pm$12.2}} \\ \\\\[0.5ex]
\textit{Mean@$5$ (\%) $\pm$ std $\uparrow$} \\
SmolLM3-3B & 6.7 {\scalebox{0.8}{$\pm$2.1}} & 20.7 {\scalebox{0.8}{$\pm$1.3}} & 10.7 {\scalebox{0.8}{$\pm$4.9}} & 14.7 {\scalebox{0.8}{$\pm$2.7}} & 4.0 {\scalebox{0.8}{$\pm$1.3}} & 13.3 {\scalebox{0.8}{$\pm$3.7}} & 14.0 {\scalebox{0.8}{$\pm$2.5}} & 12.0 {\scalebox{0.8}{$\pm$5.5}} \\
\textbf{Ours-SmolLM3-3B} & 8.0 {\scalebox{0.8}{$\pm$1.6}} & 18.0 {\scalebox{0.8}{$\pm$4.5}} & 5.3 {\scalebox{0.8}{$\pm$3.4}} & 6.7 {\scalebox{0.8}{$\pm$3.7}} & 0.0 {\scalebox{0.8}{$\pm$0.0}} & 4.7 {\scalebox{0.8}{$\pm$2.7}} & 6.7 {\scalebox{0.8}{$\pm$0.0}} & 7.0 {\scalebox{0.8}{$\pm$5.5}} \\
Qwen3-4B-Thinking-2507 & 39.3 {\scalebox{0.8}{$\pm$4.9}} & 48.7 {\scalebox{0.8}{$\pm$3.4}} & 34.0 {\scalebox{0.8}{$\pm$6.1}} & 44.0 {\scalebox{0.8}{$\pm$4.4}} & 36.0 {\scalebox{0.8}{$\pm$3.9}} & \textbf{\color{red}43.3 {\scalebox{0.8}{$\pm$3.0}}} & 42.0 {\scalebox{0.8}{$\pm$7.5}} & 41.0 {\scalebox{0.8}{$\pm$5.0}} \\
\textbf{Ours-Qwen3-4B} & \textbf{\color{red}46.0 {\scalebox{0.8}{$\pm$4.9}}} & 46.0 {\scalebox{0.8}{$\pm$3.9}} & 40.0 {\scalebox{0.8}{$\pm$2.1}} & 44.7 {\scalebox{0.8}{$\pm$4.5}} & 34.0 {\scalebox{0.8}{$\pm$7.1}} & 40.0 {\scalebox{0.8}{$\pm$3.7}} & 41.3 {\scalebox{0.8}{$\pm$4.0}} & 41.7 {\scalebox{0.8}{$\pm$4.3}} \\
Qwen-SEA-LION-v4-8B-VL & 26.0 {\scalebox{0.8}{$\pm$4.9}} & 26.7 {\scalebox{0.8}{$\pm$4.2}} & 16.7 {\scalebox{0.8}{$\pm$2.1}} & 20.0 {\scalebox{0.8}{$\pm$2.1}} & 2.7 {\scalebox{0.8}{$\pm$2.5}} & 15.3 {\scalebox{0.8}{$\pm$5.0}} & 14.7 {\scalebox{0.8}{$\pm$5.4}} & 17.4 {\scalebox{0.8}{$\pm$8.1}} \\
Qwen3-VL-8B-Thinking & 38.0 {\scalebox{0.8}{$\pm$4.0}} & \textbf{\color{red}50.0 {\scalebox{0.8}{$\pm$4.7}}} & \textbf{\color{red}41.3 {\scalebox{0.8}{$\pm$4.0}}} & \textbf{\color{red}45.3 {\scalebox{0.8}{$\pm$3.4}}} & \textbf{\color{red}38.0 {\scalebox{0.8}{$\pm$4.5}}} & 41.3 {\scalebox{0.8}{$\pm$3.4}} & \textbf{\color{red}42.7 {\scalebox{0.8}{$\pm$2.5}}} & \textbf{\color{red}42.4 {\scalebox{0.8}{$\pm$4.2}}} \\
\textbf{Ours-Qwen3-VL-8B} & 40.7 {\scalebox{0.8}{$\pm$2.5}} & 44.0 {\scalebox{0.8}{$\pm$3.9}} & 35.3 {\scalebox{0.8}{$\pm$3.4}} & 38.0 {\scalebox{0.8}{$\pm$3.4}} & 12.7 {\scalebox{0.8}{$\pm$1.3}} & 27.3 {\scalebox{0.8}{$\pm$2.5}} & 34.7 {\scalebox{0.8}{$\pm$6.2}} & 33.2 {\scalebox{0.8}{$\pm$10.5}} \\
\hline
\end{tabular*}
\caption{\label{hmmt25-main-any}
Main performance on the \texttt{HMMT25} benchmark evaluating Any-CoT,
comparing base reasoning models and OSCD post-trained models across 7 target languages (\texttt{ZH,EN,Fi,IN,TA,TH,VI}). 
}
\end{table*}

\begin{table*}[t]
\centering
\small
\setlength{\tabcolsep}{0pt} 
\begin{tabular*}{\textwidth}{l @{\extracolsep{\fill}} ccccccc >{\columncolor[gray]{0.85}}c }
\hline
& \multicolumn{8}{c}{\textbf{Target-CoT}} \\
\cline{2-9}
\textbf{Model} & \textbf{Chinese} & \textbf{English} & \textbf{Filipino} & \textbf{Indonesian} & \textbf{Tamil} & \textbf{Thai} & \textbf{Vietnamese} & \textbf{Overall} \\
\hline
\textit{Pass@$5$ (\%) $\uparrow$} \\
SmolLM3-3B & 30.0 & 60.0 & 0.0 & 0.0 & 0.0 & 0.0 & 0.0 & 12.9 {\scalebox{0.8}{$\pm$23.6}} \\
\textbf{Ours-SmolLM3-3B} & 40.0 & 53.3 & 23.3 & 36.7 & 10.0 & 16.7 & 26.7 & 29.5 {\scalebox{0.8}{$\pm$14.8}} \\
Qwen3-4B-Thinking-2507 & 73.3 & \textbf{\color{red}86.7} & 0.0 & 0.0 & 0.0 & 0.0 & 0.0 & 22.9 {\scalebox{0.8}{$\pm$39.2}} \\
\textbf{Ours-Qwen3-4B} & 73.3 & \textbf{\color{red}86.7} & \textbf{\color{red}60.0} & \textbf{\color{red}80.0} & \textbf{\color{red}46.7} & 60.0 & \textbf{\color{red}73.3} & \textbf{\color{red}68.6 {\scalebox{0.8}{$\pm$13.7}}} \\
Qwen-SEA-LION-v4-8B-VL & 36.7 & 63.3 & 16.7 & 43.3 & 26.7 & 43.3 & 40.0 & 38.6 {\scalebox{0.8}{$\pm$14.6}} \\
Qwen3-VL-8B-Thinking & 63.3 & \textbf{\color{red}86.7} & 0.0 & 0.0 & 0.0 & 0.0 & 0.0 & 21.4 {\scalebox{0.8}{$\pm$37.2}} \\
\textbf{Ours-Qwen3-VL-8B} & \textbf{\color{red}80.0} & 83.3 & 56.7 & 73.3 & 43.3 & \textbf{\color{red}70.0} & \textbf{\color{red}73.3} & \textbf{\color{red}68.6 {\scalebox{0.8}{$\pm$14.0}}} \\ \\\\[0.5ex]
\textit{Mean@$5$ (\%) $\pm$ std $\uparrow$} \\
SmolLM3-3B & 13.3 {\scalebox{0.8}{$\pm$2.1}} & 34.0 {\scalebox{0.8}{$\pm$3.3}} & 0.0 {\scalebox{0.8}{$\pm$0.0}} & 0.0 {\scalebox{0.8}{$\pm$0.0}} & 0.0 {\scalebox{0.8}{$\pm$0.0}} & 0.0 {\scalebox{0.8}{$\pm$0.0}} & 0.0 {\scalebox{0.8}{$\pm$0.0}} & 6.8 {\scalebox{0.8}{$\pm$13.0}} \\
\textbf{Ours-SmolLM3-3B} & 21.3 {\scalebox{0.8}{$\pm$4.0}} & 29.3 {\scalebox{0.8}{$\pm$6.5}} & 6.0 {\scalebox{0.8}{$\pm$5.7}} & 17.3 {\scalebox{0.8}{$\pm$2.5}} & 4.7 {\scalebox{0.8}{$\pm$1.6}} & 10.7 {\scalebox{0.8}{$\pm$3.9}} & 11.3 {\scalebox{0.8}{$\pm$3.4}} & 14.4 {\scalebox{0.8}{$\pm$8.8}} \\
Qwen3-4B-Thinking-2507 & \textbf{\color{red}52.7 {\scalebox{0.8}{$\pm$2.5}}} & \textbf{\color{red}79.3 {\scalebox{0.8}{$\pm$3.9}}} & 0.0 {\scalebox{0.8}{$\pm$0.0}} & 0.0 {\scalebox{0.8}{$\pm$0.0}} & 0.0 {\scalebox{0.8}{$\pm$0.0}} & 0.0 {\scalebox{0.8}{$\pm$0.0}} & 0.0 {\scalebox{0.8}{$\pm$0.0}} & 18.9 {\scalebox{0.8}{$\pm$33.1}} \\
\textbf{Ours-Qwen3-4B} & 42.7 {\scalebox{0.8}{$\pm$10.2}} & 70.7 {\scalebox{0.8}{$\pm$3.9}} & \textbf{\color{red}27.3 {\scalebox{0.8}{$\pm$5.7}}} & 54.0 {\scalebox{0.8}{$\pm$3.3}} & 22.7 {\scalebox{0.8}{$\pm$2.5}} & 32.7 {\scalebox{0.8}{$\pm$5.7}} & 47.3 {\scalebox{0.8}{$\pm$3.9}} & 42.5 {\scalebox{0.8}{$\pm$16.7}} \\
Qwen-SEA-LION-v4-8B-VL & 19.3 {\scalebox{0.8}{$\pm$6.5}} & 38.7 {\scalebox{0.8}{$\pm$5.4}} & 4.7 {\scalebox{0.8}{$\pm$2.7}} & 27.3 {\scalebox{0.8}{$\pm$2.5}} & 10.0 {\scalebox{0.8}{$\pm$4.7}} & 23.3 {\scalebox{0.8}{$\pm$7.6}} & 23.3 {\scalebox{0.8}{$\pm$2.1}} & 21.0 {\scalebox{0.8}{$\pm$11.2}} \\
Qwen3-VL-8B-Thinking & 50.0 {\scalebox{0.8}{$\pm$2.1}} & 74.7 {\scalebox{0.8}{$\pm$3.4}} & 0.0 {\scalebox{0.8}{$\pm$0.0}} & 0.0 {\scalebox{0.8}{$\pm$0.0}} & 0.0 {\scalebox{0.8}{$\pm$0.0}} & 0.0 {\scalebox{0.8}{$\pm$0.0}} & 0.0 {\scalebox{0.8}{$\pm$0.0}} & 17.8 {\scalebox{0.8}{$\pm$31.2}} \\
\textbf{Ours-Qwen3-VL-8B} & 50.0 {\scalebox{0.8}{$\pm$4.7}} & 74.0 {\scalebox{0.8}{$\pm$4.9}} & 25.3 {\scalebox{0.8}{$\pm$5.4}} & \textbf{\color{red}58.0 {\scalebox{0.8}{$\pm$5.4}}} & \textbf{\color{red}23.3 {\scalebox{0.8}{$\pm$4.7}}} & \textbf{\color{red}53.3 {\scalebox{0.8}{$\pm$3.0}}} & \textbf{\color{red}48.7 {\scalebox{0.8}{$\pm$5.0}}} & \textbf{\color{red}47.5 {\scalebox{0.8}{$\pm$17.9}}} \\
\hline
\end{tabular*}
\caption{\label{aime25-main-target}
Main performance on the \texttt{AIME25} benchmark evaluating Target-CoT,
comparing base reasoning models and OSCD post-trained models across 7 target languages (\texttt{ZH,EN,Fi,IN,TA,TH,VI}). 
}
\end{table*}

\begin{table*}[t]
\centering
\small
\setlength{\tabcolsep}{0pt} 
\begin{tabular*}{\textwidth}{l @{\extracolsep{\fill}} ccccccc >{\columncolor[gray]{0.85}}c }
\hline
& \multicolumn{8}{c}{\textbf{Target-CoT}} \\
\cline{2-9}
\textbf{Model} & \textbf{Chinese} & \textbf{English} & \textbf{Filipino} & \textbf{Indonesian} & \textbf{Tamil} & \textbf{Thai} & \textbf{Vietnamese} & \textbf{Overall} \\
\hline
\textit{Pass@$5$ (\%) $\uparrow$} \\
SmolLM3-3B & 16.7 & 33.3 & 0.0 & 0.0 & 0.0 & 0.0 & 0.0 & 7.1 {\scalebox{0.8}{$\pm$13.1}} \\
\textbf{Ours-SmolLM3-3B} & 13.3 & 30.0 & 10.0 & 13.3 & 0.0 & 6.7 & 13.3 & 12.4 {\scalebox{0.8}{$\pm$9.2}} \\
Qwen3-4B-Thinking-2507 & \textbf{\color{red}50.0} & 56.7 & 0.0 & 0.0 & 0.0 & 0.0 & 0.0 & 15.2 {\scalebox{0.8}{$\pm$26.1}} \\
\textbf{Ours-Qwen3-4B} & 46.7 & 56.7 & 26.7 & \textbf{\color{red}53.3} & \textbf{\color{red}23.3} & 30.0 & \textbf{\color{red}50.0} & 41.0 {\scalebox{0.8}{$\pm$13.8}} \\
Qwen-SEA-LION-v4-8B-VL & 33.3 & 40.0 & 6.7 & 23.3 & 6.7 & 26.7 & 23.3 & 22.9 {\scalebox{0.8}{$\pm$12.5}} \\
Qwen3-VL-8B-Thinking & 40.0 & 56.7 & 0.0 & 0.0 & 0.0 & 0.0 & 0.0 & 13.8 {\scalebox{0.8}{$\pm$24.1}} \\
\textbf{Ours-Qwen3-VL-8B} & 43.3 & \textbf{\color{red}60.0} & \textbf{\color{red}33.3} & 46.7 & 16.7 & \textbf{\color{red}40.0} & \textbf{\color{red}50.0} & \textbf{\color{red}41.4 {\scalebox{0.8}{$\pm$13.7}}} \\ \\\\[0.5ex]
\textit{Mean@$5$ (\%) $\pm$ std $\uparrow$} \\
SmolLM3-3B & 6.0 {\scalebox{0.8}{$\pm$3.3}} & 20.7 {\scalebox{0.8}{$\pm$1.3}} & 0.0 {\scalebox{0.8}{$\pm$0.0}} & 0.0 {\scalebox{0.8}{$\pm$0.0}} & 0.0 {\scalebox{0.8}{$\pm$0.0}} & 0.0 {\scalebox{0.8}{$\pm$0.0}} & 0.0 {\scalebox{0.8}{$\pm$0.0}} & 3.8 {\scalebox{0.8}{$\pm$7.8}} \\
\textbf{Ours-SmolLM3-3B} & 6.7 {\scalebox{0.8}{$\pm$2.1}} & 15.3 {\scalebox{0.8}{$\pm$5.4}} & 4.0 {\scalebox{0.8}{$\pm$3.9}} & 6.7 {\scalebox{0.8}{$\pm$3.7}} & 0.0 {\scalebox{0.8}{$\pm$0.0}} & 4.7 {\scalebox{0.8}{$\pm$2.7}} & 5.3 {\scalebox{0.8}{$\pm$1.6}} & 6.1 {\scalebox{0.8}{$\pm$4.7}} \\
Qwen3-4B-Thinking-2507 & \textbf{\color{red}30.0 {\scalebox{0.8}{$\pm$3.0}}} & 46.0 {\scalebox{0.8}{$\pm$2.5}} & 0.0 {\scalebox{0.8}{$\pm$0.0}} & 0.0 {\scalebox{0.8}{$\pm$0.0}} & 0.0 {\scalebox{0.8}{$\pm$0.0}} & 0.0 {\scalebox{0.8}{$\pm$0.0}} & 0.0 {\scalebox{0.8}{$\pm$0.0}} & 10.9 {\scalebox{0.8}{$\pm$19.1}} \\
\textbf{Ours-Qwen3-4B} & 20.0 {\scalebox{0.8}{$\pm$6.0}} & 44.0 {\scalebox{0.8}{$\pm$3.9}} & \textbf{\color{red}11.3 {\scalebox{0.8}{$\pm$6.2}}} & \textbf{\color{red}36.0 {\scalebox{0.8}{$\pm$3.9}}} & 10.0 {\scalebox{0.8}{$\pm$3.7}} & 15.3 {\scalebox{0.8}{$\pm$3.4}} & \textbf{\color{red}27.3 {\scalebox{0.8}{$\pm$5.7}}} & 23.4 {\scalebox{0.8}{$\pm$12.9}} \\
Qwen-SEA-LION-v4-8B-VL & 14.7 {\scalebox{0.8}{$\pm$3.4}} & 23.3 {\scalebox{0.8}{$\pm$5.6}} & 1.3 {\scalebox{0.8}{$\pm$1.6}} & 15.3 {\scalebox{0.8}{$\pm$1.6}} & 1.3 {\scalebox{0.8}{$\pm$1.6}} & 10.7 {\scalebox{0.8}{$\pm$6.8}} & 7.3 {\scalebox{0.8}{$\pm$3.9}} & 10.6 {\scalebox{0.8}{$\pm$8.0}} \\
Qwen3-VL-8B-Thinking & 28.0 {\scalebox{0.8}{$\pm$4.5}} & \textbf{\color{red}50.0 {\scalebox{0.8}{$\pm$4.7}}} & 0.0 {\scalebox{0.8}{$\pm$0.0}} & 0.0 {\scalebox{0.8}{$\pm$0.0}} & 0.0 {\scalebox{0.8}{$\pm$0.0}} & 0.0 {\scalebox{0.8}{$\pm$0.0}} & 0.0 {\scalebox{0.8}{$\pm$0.0}} & 11.1 {\scalebox{0.8}{$\pm$20.1}} \\
\textbf{Ours-Qwen3-VL-8B} & \textbf{\color{red}30.0 {\scalebox{0.8}{$\pm$3.7}}} & 44.0 {\scalebox{0.8}{$\pm$3.9}} & \textbf{\color{red}11.3 {\scalebox{0.8}{$\pm$4.5}}} & 31.3 {\scalebox{0.8}{$\pm$3.4}} & \textbf{\color{red}10.7 {\scalebox{0.8}{$\pm$2.5}}} & \textbf{\color{red}25.3 {\scalebox{0.8}{$\pm$3.4}}} & 26.7 {\scalebox{0.8}{$\pm$4.7}} & \textbf{\color{red}25.6 {\scalebox{0.8}{$\pm$11.7}}} \\
\hline
\end{tabular*}
\caption{\label{hmmt25-main-target}
Main performance on the \texttt{HMMT25} benchmark evaluating Target-CoT,
comparing base reasoning models and OSCD post-trained models across 7 target languages (\texttt{ZH,EN,Fi,IN,TA,TH,VI}). 
}
\end{table*}

\clearpage 

\end{document}